\documentclass[11pt]{article}

\usepackage{bbm}
\usepackage{stmaryrd}

\usepackage{algorithmic}

\usepackage[normalem]{ulem} 
\newcommand\hl{\bgroup\markoverwith
  {\textcolor{yellow}{\rule[-.5ex]{2pt}{2.5ex}}}\ULon}

\usepackage{epstopdf}

\usepackage{fullpage}
\usepackage[ruled, linesnumbered, vlined, commentsnumbered]{algorithm2e}
\usepackage{graphicx,subfig} 
\usepackage{amsfonts,amssymb,amsmath,amsthm,amsopn}	
\usepackage{hhline,booktabs,colortbl,multirow,tabularx,diagbox,threeparttable} 

\usepackage{enumerate}
\usepackage{authblk}
\usepackage{footnote}
\usepackage{hyperref}
\usepackage{prettyref}
\usepackage{cite}
\usepackage{setspace}
\usepackage{color}
\usepackage{xcolor}  
\usepackage{scrpage2}
\usepackage{lscape}

\usepackage{tikz}
\usepackage{pgfplots}
\usetikzlibrary{positioning,shapes,shadows,arrows,calc}
\tikzstyle{component}=[rectangle, draw=black, rounded corners, fill=blue!40, drop shadow, text centered, anchor=north, text=white, minimum height=1cm]
\tikzstyle{arrow}=[->, thick]

\pgfplotsset{compat=1.12}
\usetikzlibrary{intersections}
\usetikzlibrary{pgfplots.statistics}
\usepgfplotslibrary{fillbetween}

\definecolor{myblue}{RGB}{34,31,217}
\definecolor{mycyan}{gray}{.7}
\definecolor{Gray}{gray}{0.9}

\DeclareMathOperator*{\argmax}{argmax}
\DeclareMathOperator*{\argmin}{argmin}

\newrefformat{fig}{Fig.~\ref{#1}}
\newrefformat{tab}{Table~\ref{#1}}
\newrefformat{sec}{Section~\ref{#1}}
\newrefformat{app}{Appendix~\ref{#1}}
\newrefformat{alg}{Algorithm~\ref{#1}}
\newrefformat{property}{Property~\ref{#1}}
\newrefformat{theorem}{Theorem~\ref{#1}}
\newrefformat{corollary}{Corollary~\ref{#1}}
\newrefformat{proposition}{Proposition~\ref{#1}}
\newrefformat{def}{Definition~\ref{#1}}
\newrefformat{eq}{equation~(\ref{#1})}

\title{\vspace{-1ex}\LARGE\textbf{Bayesian Network Based Label Correlation Analysis For Multi-label Classifier Chain}\footnote{This manuscript is currently under peer review for possible publication. The reviewer can use this version interchangeably.}}

\author[1,2]{\normalsize Ran Wang}
\author[1,2]{\normalsize Suhe Ye}
\author[3]{\normalsize Ke Li$^+$}
\author[4]{\normalsize Sam Kwong}

\affil[1]{College of Mathematics and Statistics, Shenzhen University, Shenzhen 518060, China.}
\affil[2]{Shenzhen Key Laboratory of Advanced Machine Learning and Applications, Shenzhen University, Shenzhen 518060, China.}
\affil[3]{\normalsize Department of Computer Science, University of Exeter, EX4 4QF, Exeter, UK}
\affil[4]{Department of Computer Science, City University of Hong Kong, 83 Tat Chee Avenue, Kowloon, Hong Kong.}
\affil[$^+$]{\normalsize Email: \texttt{k.li@exeter.ac.uk}}

\date{}

\begin{document}
\maketitle

\vspace{-3ex}
{\normalsize\textbf{Abstract: } Classifier chain (CC) is a multi-label learning approach that constructs a sequence of binary classifiers according to a label order. Each classifier in the sequence is responsible for predicting the relevance of one label. When training the classifier for a label, proceeding labels will be taken as extended features. If the extended features are highly correlated to the label, the performance will be improved, otherwise, the performance will not be influenced or even degraded. How to discover label correlation and determine the label order is critical for CC approach. This paper employs Bayesian network (BN) to model the label correlations and proposes a new BN-based CC method (BNCC). First, conditional entropy is used to describe the dependency relations among labels. Then, a BN is built up by taking nodes as labels and weights of edges as their dependency relations. A new scoring function is proposed to evaluate a BN structure, and a heuristic algorithm is introduced to optimize the BN. At last, by applying topological sorting on the nodes of the optimized BN, the label order for constructing CC model is derived. Experimental comparisons demonstrate the feasibility and effectiveness of the proposed method.}


\section{Introduction}\label{sec.introduction}

Multi-label learning (MLL) deals with the problems in which an instance can be assigned to multiple classes simultaneously~\cite{madjarov2012extension,zhang2014review}. Given a label set $\mathbb{L}=\{l_1,l_2,\ldots,l_M\}$, traditional single-label learning (SLL)~\cite{wang2016ambiguity,wang2018incorporating} constructs a model that maps the instances from the feature space to the discrete label set, i.e., $h:\ \mathbf{x}\rightarrow\mathbb{L}$, while MLL constructs a model that maps the instances from the feature space to the powerset of the label set, i.e., $h:\ \mathbf{x}\rightarrow2^{\mathbb{L}}$. In recent decades, MLL has been extensively studied and has been applied to a wide range of application domains like text classification~\cite{ueda2003parametric,yu2005multi,zhang2006multilabel,yang2009effective}, image recognition~\cite{boutella2004learning,zhou2007multiinstance,qi2009two,sun2014multilabel,wu2017active}, and music categorization~\cite{trohidis2008multilabel}, etc.

\begin{figure}[t]\centering
\includegraphics[width=2.5in]{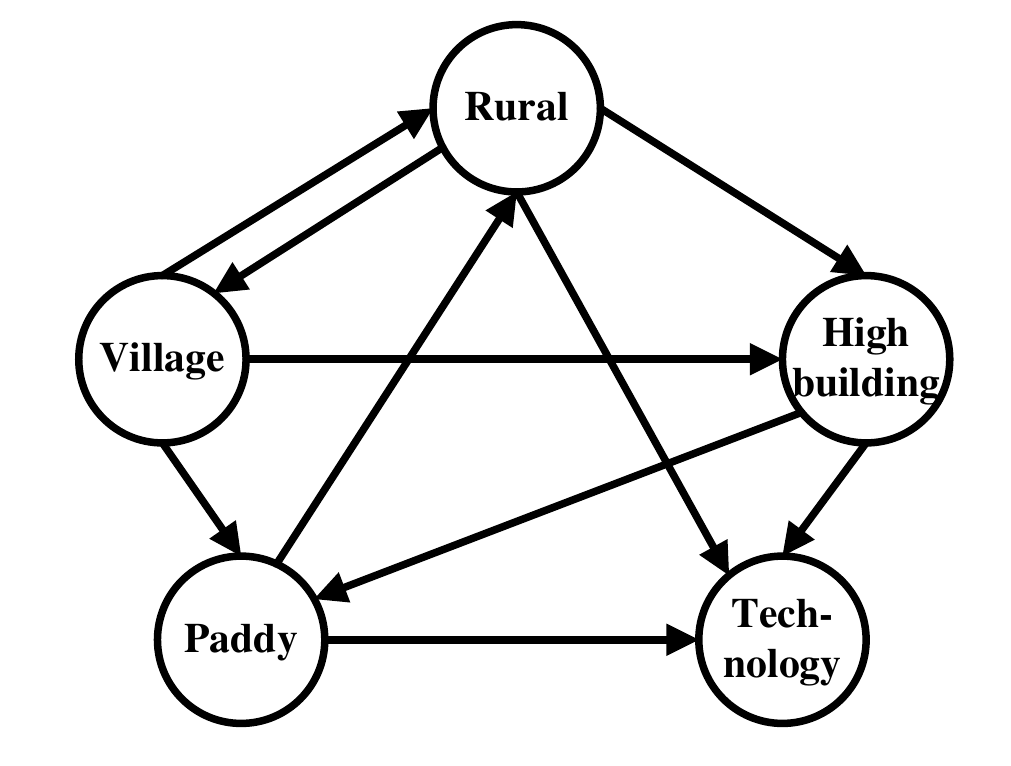}
\caption{Directed network composed of five labels.}
\label{fig.DAG}
\end{figure}

A straightforward solution to MLL is the label powerset (LP) method. It transforms the original multi-label problem into a single-label problem by treating each element in $2^\mathbb{L}$ as a single class. However, the complexity of LP method is extremely high since the number of classes grows exponentially with the increase of $|\mathbb{L}|$. It is a big challenge to train effective MLL models with a reasonable time complexity. In general, three groups of approximate methods have been proposed, i.e., problem transformation methods (PTMs)~\cite{zhang2014review}, ensemble methods (EMs)~\cite{Rokach2014Ensemble,Moyano2018Review}, and algorithm adaption methods (AAMs)~\cite{zhang2007mlknn,guo2011multi,guo2011adaptive}. Among them, PTMs are the most efficient by decomposing the multi-label problem into a set of smaller single-label problems in either binary or multi-class case.  The most fundamental PTMs include binary relevance (BR)~\cite{boutella2004learning} and calibrated label ranking (CLR)~\cite{Furnkranz2008multilabel}. BR trains a binary classifier for each label independently, while CLR trains a binary classifier for each pair of the labels. These two methods are easy to implement with a relatively low time complexity, but they ignore the mutual influences among labels~\cite{zhang2012batch,huang2012multilabel,zhang2014multilabel} that may affect the final performance. For example, given a five-label image recognition problem with $\mathbb{L}=\{{\rm Village}, {\rm Rural}, {\rm Paddy}, {\rm High\ building}, {\rm Technology}\}$. There may exist certain relationships indicating that the decision of one label (denoted as $l_{\rm s}\in\mathbb{L}$) has an influence on the decision of another label (denoted as $l_{\rm e}\in\mathbb{L}$). If we treat each label as a node and use directed edges to link related nodes, e.g., $l_{\rm s}\rightarrow l_{\rm e}$, then a directed network connecting all the labels can be constructed as shown in Fig.~\ref{fig.DAG}. Discovering and incorporating such label correlations can help constructing a better MLL model.

\begin{figure}[t]
\centering
\subfloat[Positive relationship between {\it Rural}-{\it Village} or {\it Rural-Paddy}]{\includegraphics[width=5in]{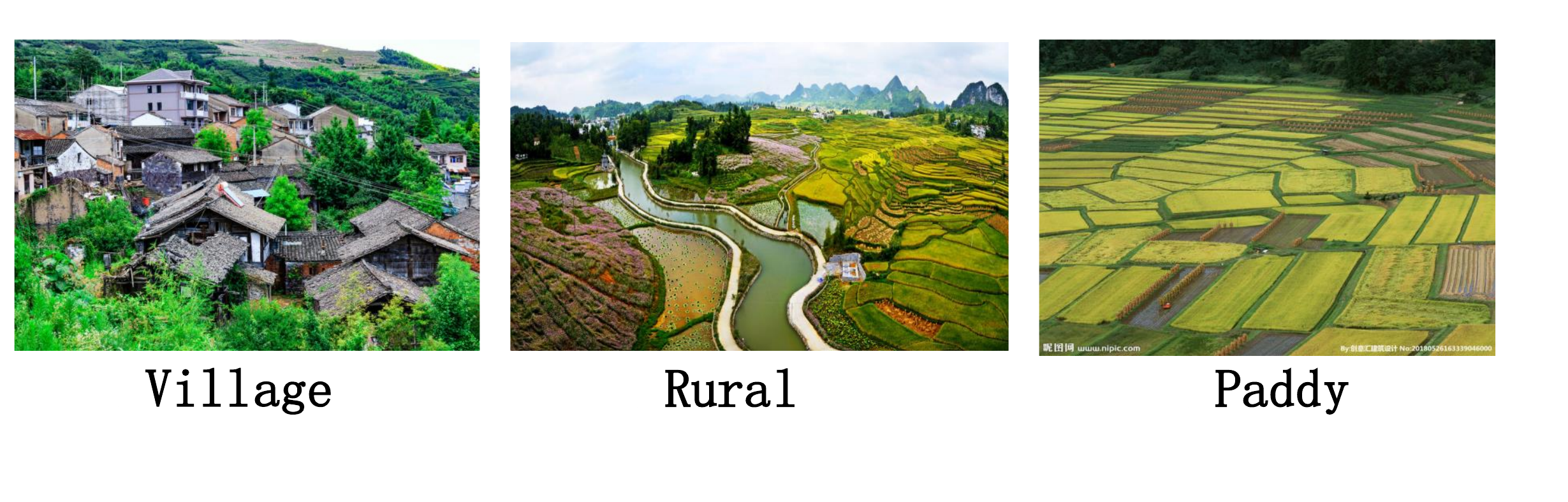}}
\subfloat[Negative relationship between {\it Rural}-{\it High building} or {\it Rural}-{\it Technology}]{\includegraphics[width=5.2in]{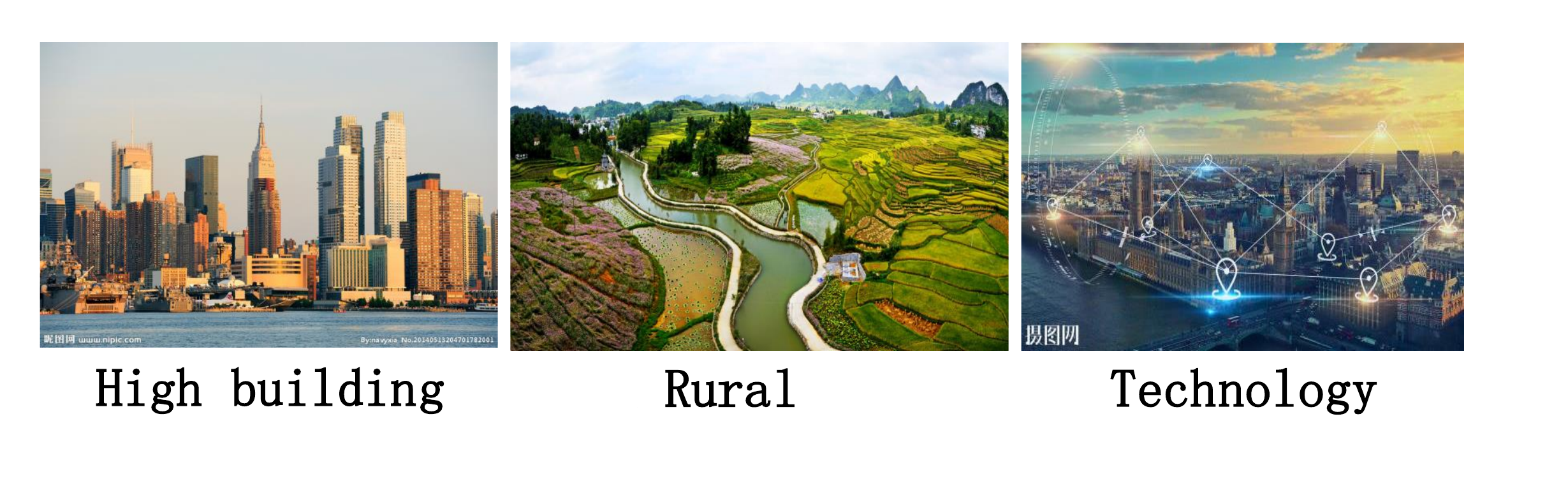}}
\caption{Relationships among labels.}
\label{fig.labelRelation}
\end{figure}

Usually, there are two kinds of relationships among labels, i.e., positive relationship and negative relationship. Positive relationship refers to the co-occurrence or co-disappearance of labels, as shown in Fig.~\ref{fig.labelRelation}(a), when ${\rm Rural}$ appears in an image, ${\rm Village}$ or ${\rm Paddy}$ is also likely to appear; while negative relationship refers to the mutually exclusive relations of labels, as shown in Fig.~\ref{fig.labelRelation}(b), when ${\rm Rural}$ appears, ${\rm High\ building}$ or ${\rm Technology}$ is unlikely to appear. Both positive and negative relationships are useful for modeling the label correlations.

Classifier chain (CC) is a PTM that tries to make use of the label correlations~\cite{read2009classifier}. Similar to BR, CC constructs $|\mathbb{L}|$ binary classifiers and each classifier is responsible for predicting the relevance of one label. However, the classifiers are trained sequentially by following a pre-defined label order, and the input feature vector for one label is extended by the labels ordered before it. The key issue in CC approach is to find the optimal label order. If the predecessors of a label are highly correlated to it, then the extended features can help improve the performance of the corresponding classifier, otherwise not. Original CC approach determines the label order randomly, which has a risk of low performance and low robustness. Later, ensemble CC (ECC) was proposed~\cite{read2011classifier}. It trains multiple CCs based on different random orders, and gets the final prediction by collective voting. ECC can reduce the risk of low performance, but the time complexity is high. Another scheme called double-Monte Carlo CC (M2CC) was proposed~\cite{read2013efficient}, which models the dependencies of labels based on their co-occurrence. It finds the possible chain-sequences during training stage and determines the best chain by efficient inference through Monte Carlo optimization. Furthermore, group sensitive CC (GCC) was proposed~\cite{huang2015group} by considering local and positive label correlations. To assume that similar instances tend to have similar labels, GCC clusters the instances into groups. It builds a CC for each group and predicts an unseen instance by the CC trained on its nearest group. Moreover, enhanced CC with $k$-means clustering algorithm ($k$m-CC) was proposed~\cite{yu2015improved}. By employing $k$-means algorithm several times, correlations among labels are discovered and the order of classifiers is determined. It is noteworthy that the above-introduced methods can help improve the performance of CC approach, but the time complexity is usually high. Besides, most of them analyze label correlations based on co-occurrence, whereas the mutually exclusive relations are neglected. According to these disadvantage, a comprehensive model is desired for label correlation analysis

Bayesian network (BN), known as a directed acyclic graph (DAG), is a probabilistic graphical model that learns the properties of a set of random variables and their conditional probability distributions~\cite{pearl1991probabilistic}. BN has a wide range of applications such as data mining~\cite{naili2019stability}, fault detection~\cite{amin2018fault}, safety analysis~\cite{zarei2019safety}, agricultural research~\cite{drury2017survey}, bioinformatics~\cite{tamada2011estimating,li2016max}, and so on. In general, nodes in BN represent random variables, and edges connecting two nodes represent the relations between variables. If there is no edge connecting two nodes, then the two random variables are independent of each other. Conversely, if two nodes are connected by an edge, then the {\it parent} node (i.e., start-point of the edge) and the {\it child} node (i.e., end-point of the edge) are causally or non-conditionally dependent, which will generate a conditional probability value. By imposing a BN constraint on the random order, an improved CC approach with tree-based structure was proposed~\cite{sucar2014multi}. Furthermore, BN-augmented naive Bayes classifiers are used as the base models for CC approach~\cite{varando2016decision}. However, to the best of our knowledge, using BN model for comprehensive label correlation analysis has not been investigated yet, which will be the main focus of this paper. By further introducing a fast label ordering method, a new BN-based CC approach (BNCC) is proposed. The contributions of this paper are listed as follows:
\begin{itemize}
\item Conditional entropy is used to model the dependency degree of a label on other labels, which makes use of both positive relationship and negative relationship. A fully connected directed cyclic graph (DCG) is built up as the initial structure, where nodes represent labels and weights of edges indicate the dependency degrees among connected labels.
\item An algorithm is proposed to refine a DCG into a DAG by breaking cycles iteratively, which guarantees to generate effective BN structure. We also propose to use topological sorting on the nodes of a DAG, in order to get efficacious label order from a BN structure.
\item A new scoring function is proposed to evaluate the quality of BN, which includes the dependency degree calculated by conditional entropy and a complexity penalization term. Since learning the optimal BN is inference intractable, a heuristic algorithm is proposed to get approximate solutions based on the scoring function.
\item We conduct extensive experimental comparisons between the proposed method and several state-of-the-art MLL approaches. Empirical studies show that the proposed method can generate effective CC model with a relatively low time complexity in both training and testing.
\end{itemize}

The remainder of this paper is organized as follows. In section~\ref{sec.backgroud}, we introduce some background knowledge. In section~\ref{sec.proposedModel}, we present our proposed method. In section~\ref{sec.experiment}, extensive experimental comparisons are conducted to show the advantages of the proposed method. Finally, conclusions are given in section~\ref{sec.conlusions}.

\section{Background Knowledge}\label{sec.backgroud}

In this section, we will present some background knowledge on CC approach, label correlation, and BN model.

\subsection{CC Approach}\label{subsec.CC}

Suppose that $\mathcal{X}=\mathbf{R}^d$ is a $d$-dimensional instance space, $\mathbb{L}=\{l_1,\ldots,l_M\}$ is a label set, and $\mathcal{Y}=\{0,1\}^M$ is a decision space with regard to $\mathbb{L}$. We denote by $\mathbb{D}=\{(\mathbf{x}_i,\mathbf{y}_i)\}_{i=1}^{N}$ as the training set with $N$ instances, where $\mathbf{x}_i=[x_{i1},x_{i2},\ldots ,x_{id}]\in\mathcal{X}$ is the feature vector for the $i$-th instance and $\mathbf{y}_i=[y_{i1},y_{i2},\ldots,y_{iM}]\in\mathcal{Y}$ is the label vector of $\mathbf{x}_i$. We have $y_{ij}=1$ if $\mathbf{x}_i$ has the $j$-th label and $y_{ij}=0$ otherwise, where $j=1,2,\ldots,M$. MLL aims to train a function $h: \mathcal{X}\rightarrow\mathcal{Y}$ from $\mathbb{D}$, such that $h$ can predict the label vector of unseen instance $\hat{\mathbf{x}}$, i.e., $\hat{\mathbf{y}}=h(\hat{\mathbf{x}})$.

CC approach decomposes the $M$-label problem into a chain of $M$ binary problems. One classifier in the chain deals with a binary problem regarding a label in $\mathbb{L}$. In the training phase, the binary classifiers are constructed following a pre-defined order, where the input feature vector for the current classifier is extended by the previous labels. In specific, CC approach firstly generates a new order of the labels by a permutation function $\tau$, i.e.,

\begin{displaymath}
\tau:\{1,\ldots,M\}\rightarrow\{1,\ldots,M\}.
\end{displaymath}
The training set for $l_{\tau(1)}$ is $\mathbb{D}_{1}=\{(\mathbf{x}_i,{y}_{i\tau(1)})\}_{i=1}^{N}$, and the first classifier is constructed as $h_1: \mathbf{R}^d\rightarrow\{0,1\}$. Then, the training set for $l_{\tau(2)}$ is extended as $\mathbb{D}_{2}=\{([\mathbf{x}_i,{y}_{i\tau(1)}],{y}_{i\tau(2)})\}_{i=1}^{N}$, and the second classifier is constructed as $h_2: \mathbf{R}^{d+1}\rightarrow\{0,1\}$. Following this rule, the general form of the training set for $l_{\tau(j)}$ is

\begin{displaymath}
\mathbb{D}_{j}=\{([\mathbf{x}_i,{\rm\bf Pre}_{i\tau(j)}],{y}_{i\tau(j)})\}_{i=1}^{N},
\end{displaymath}
where ${\rm\bf Pre}_{i\tau(j)}=[{y}_{i\tau(1)},{y}_{i\tau(2)},\ldots,{y}_{i\tau(j-1)}]$, and ${\rm\bf Pre}_{i\tau(1)}=\emptyset$. As a result, the $j$-th classifier is constructed as

\begin{displaymath}
h_j: \mathbf{R}^{d+j-1}\rightarrow\{0,1\},
\end{displaymath}
where $j=1,2,\ldots,M$.

In the testing phase, given an unseen instance $\hat{\mathbf{x}}$, the decision for $\l_{\tau(1)}$ is predicted as $\hat{y}_{\tau(1)}=h_1(\hat{\mathbf{x}})$. The decision for $l_{\tau(2)}$ is predicted as $\hat{y}_{\tau(2)}=h_2([\hat{\mathbf{x}},h_1(\hat{\mathbf{x}})])$. Following this rule, the decision for $l_{\tau(j)}$ is predicted as

\begin{displaymath}
\hat{y}_{\tau(j)}=h_j([\hat{\mathbf{x}},\hat{{\rm\bf Pre}}_{\tau(j)}]),
\end{displaymath}
where $\hat{{\rm\bf Pre}}_{\tau(j)}=[\hat{{\rm\bf Pre}}_{\tau(j-1)},h_{j-1}([\hat{\mathbf{x}},\hat{{\rm\bf Pre}}_{\tau(j-1)}])]$, $\hat{{\rm\bf Pre}}_{\tau(1)}=\emptyset$, and $j=1$, $2$, $\ldots$, $M$.

Table~\ref{tab.ccModel} shows the extended feature vectors of a training instance $(\mathbf{x},\mathbf{y})=([0.8,2.5,0.6,1.3,4.2],[1,0,1,0,0,1])$ in a six-label problem. Suppose that the label order is determined as $\tau: [1,2,3,4,5,6]\rightarrow[2,4,1,6,5,3]$, then six classifiers are trained by strictly following the order $\tau$, where $h_j$ is responsible for predicting the relevance of $l_{\tau(j)}$, $j=1,2,\ldots,6$.

In CC approach, if a label is strongly correlated to its predecessors, the extended feature dimensions can help improve the performance of the corresponding classifier, otherwise, the result will not be influenced or even degraded. That is, the performance highly relies on the label order $\tau$. How to get the optimal $\tau$ is still an ongoing research topic.

\begin{table}[t]
\caption{Extended Feature Vectors for CC Approach}
\begin{center}
\scalebox{0.9}
{\begin{tabular}{llll}
\hline
$j$ & $l_{\tau(j)}$ & Feature vector $[\mathbf{x},{\rm\bf Pre}_{\tau(j)}]$ for training $h_j$ &$y_{\tau(i)}$ \\
\hline
$1$ & $l_2$ & $[0.8,2.5,0.6,1.3,4.2]$            & 0 \\
$2$ & $l_4$ & $[0.8,2.5,0.6,1.3,4.2,0]$          & 0 \\
$3$ & $l_1$ & $[0.8,2.5,0.6,1.3,4.2,0,0]$        & 1 \\
$4$ & $l_6$ & $[0.8,2.5,0.6,1.3,4.2,0,0,1]$      & 1 \\
$5$ & $l_5$ & $[0.8,2.5,0.6,1.3,4.2,0,0,1,1]$    & 0 \\
$6$ & $l_3$ & $[0.8,2.5,0.6,1.3,4.2,0,0,1,1,0]$  & 1 \\
\hline\noalign{\smallskip}
\end{tabular}}
\label{tab.ccModel}
\end{center}
\end{table}

\subsection{Correlations Among Labels}

Given a $M$-label problem with $\mathbb{L}=\{l_1,l_2,\dots,l_M\}$, we use $l_j$ to denote the name of the $j$-th label and $y_j\in\{0,1\}$ to denote a specific value that $l_j$ can take. Considering any pair of distinct labels $l_j,l_k\in\mathbb{L}$ ($l_j\neq l_k$), it is known that $p(l_j=y_j,l_k=y_k)$ is the joint probability, $p(l_j=y_j)$ and $p(l_k=y_k)$ are the marginal probabilities, $p(l_j=y_j|l_k=y_k)$ and $p(l_k=y_k|l_j=y_j)$ are the conditional probabilities. We denote $p(l_j=y_j,l_k=y_k)$, $p(l_j=y_j)$, $p(l_k=y_k)$, $p(l_j=y_j|l_k=y_k)$ and $p(l_k=y_k|l_j=y_j)$ as $p(y_j,y_k)$, $p_{l_j}(y_j)$, $p_{l_k}(y_k)$, $p_{l_j}(y_j|y_k)$ and $p_{l_k}(y_k|y_j)$ for short. Then, the following remarks hold.

\begin{enumerate}
\item~$p_{l_j}(y_j)=\sum\nolimits_{y_k\in\{0,1\}}p(y_j,y_k)$ and $p_{l_k}(y_k)=\sum\nolimits_{y_j\in\{0,1\}}p(y_j,y_k)$.
\item~$\sum\nolimits_{y_j\in\{0,1\}}p_{l_j}(y_j)=\sum\nolimits_{y_k\in\{0,1\}}p_{l_k}(y_k)=1$.
\item~$p(y_j,y_k)=p_{l_j}(y_j)\times p_{l_k}(y_k|y_j)=p_{l_k}(y_k)\times p_{l_j}(y_j|y_k)$.
\end{enumerate}
Based on these remarks, the following definition is given.

\newtheorem{mydef}{Definition}

\begin{mydef}\label{def.labelIndep}
(Label Independence) Given a pair of labels $(l_j,l_k)$ and their probability distributions, $l_j$ and $l_k$ are considered as independent if and only if they satisfy

\begin{equation}\label{eq.labelIndep}
p(y_j,y_k)=p_{l_j}(y_j)\times p_{l_k}(y_k)
\end{equation}
for any $y_j,y_k\in\{0,1\}$.
\end{mydef}

From Definition~\ref{def.labelIndep} and Remark~3, it is easy to know that if $l_j$ and $l_k$ are independent, the marginal probability equals to conditional probability, i.e., $p_{l_j}(y_j)=p_{l_j}(y_j|y_k)$ and $p_{l_k}(y_k)=p_{l_k}(y_k|y_j)$. Let us consider a specific example in Table~\ref{tab.proDistribution}, where the probabilities are calculated from an artificial data set. We can obtain $p(0,0)=0.4$ and $p_{l_1}(0)\times p_{l_2}(0)=0.4\times 0.7=0.28$, thus we have $p(0,0)\neq p_{l_1}(0)\times p_{l_2}(0)$. This inequality also holds for $p(0,1)$, $p(1,0)$, and $p(1,1)$. In this case, $l_1$ and $l_2$ are not independent, and this phenomenon exits in almost all multi-label data sets.

\begin{table}[t]
\caption{An Example for Label Probabilities}
\begin{center}
\scalebox{0.9}
{\begin{tabular}{c|cc|c}
\hline
$p(y_1,y_2)$        & $y_2=0$ & $y_2=1$ & $p_{l_1}(y_1)$ \\
\hline
$y_1=0$             & $0.4$ & $0.0$ & $0.4$ \\
$y_1=1$             & $0.3$ & $0.3$ & $0.6$ \\
\hline
$p_{l_2}(y_2)$      & $0.7$ & $0.3$ & 1 \\
\hline\noalign{\smallskip}
\end{tabular}}
\label{tab.proDistribution}
\end{center}
\end{table}

In classical correlation analysis, we know that if two variables are independent, then they are linearly uncorrelated. Conversely, if two variables are not independent, linear correlation may exist between them. However, in MLL, traditional linear correlation coefficient may not be a good measure due to the binary value configuration for each label. Designing a meaningful correlation model is a key issue in this work.

Moreover, most existing works for label correlation analysis only consider the co-occurrence of two labels. In this paper, we define the co-occurrence and co-disappearance of two labels, i.e., $p(1,1)$ and $p(0,0)$, as positive relationships, and the mutually exclusive relations, i.e., $p(0,1)$ and $p(1,0)$, as negative relationships, which will be utilized together to improve CC approach.

\subsection{BN Model}

BN is a probabilistic graphical model that captures the dependency relations among a set of variables by a DAG. A BN is denoted by $\mathcal{B}=(\mathbb{G},\Theta)$, which consists of two components, i.e., $\mathbb{G}$ and $\Theta$. The first component $\mathbb{G}=(\mathbb{V},\mathbb{E})$ is a DAG, where $\mathbb{V}$ is a set of nodes and $\mathbb{E}$ is a set of directed edges connecting the nodes. Usually, each node represents a random variable and each edge represents a directed dependency relation between two variables. The second component $\Theta$ is a set of parameters that measures the network, which is described by a set of conditional probabilities.

In this paper, we discuss BN under MLL scenario, thus nodes represent labels and edges reflect label correlations. According to the classical chain rule, a unique joint probability distribution over $\mathbb{L}$ is given by

\begin{equation}\label{eq.BN}
\centering
\begin{split}
p(y_1,\dots,y_M)=p(y_1)p(y_2|y_1)p(y_3|y_1,y_2)\\
\dots p(y_M|y_1,y_2,\ldots,y_{M-1}).
\end{split}
\end{equation}
Considering the conditional probabilities in Eq.~(\ref{eq.BN}), we have the following definition.

\begin{mydef}\label{def.conditionalIndep}
(Conditional Independence) Let $p$ be a joint probability distribution over the labels in $\mathbb{L}$, and $\{l_j,l_k,l_q\}$ be a subset of $\mathbb{L}$. If $p(y_j|y_k,y_q)=p(y_j|y_q)$ holds for all possible combinations of $y_j,y_k,y_q\in\{0,1\}$, then we say that $l_j$ and $l_k$ are conditionally independent given $l_q$.
\end{mydef}

\begin{figure}[t]
\centering
\subfloat[Tail-to-Tail]{\includegraphics[width=1.0in]{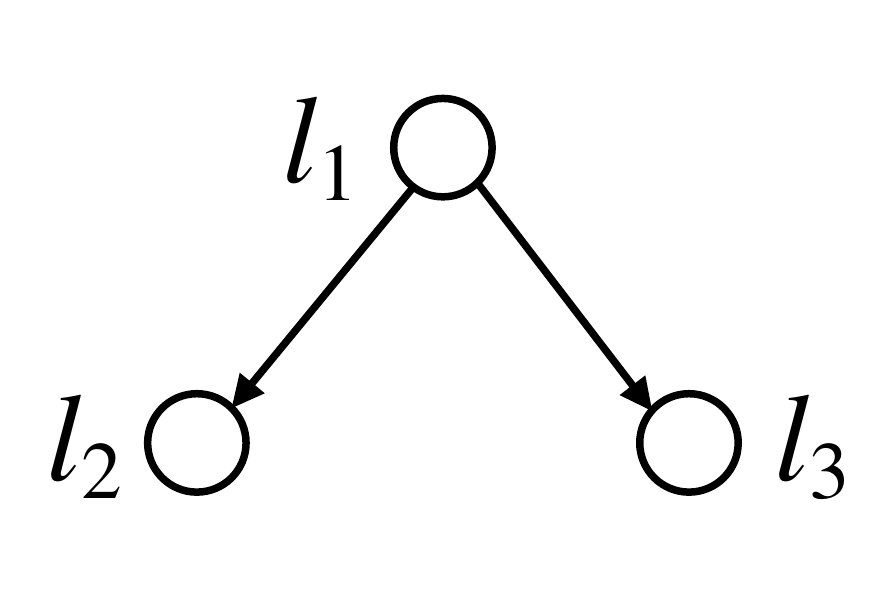}}\quad
\subfloat[Head-to-Tail]{\includegraphics[width=1.0in]{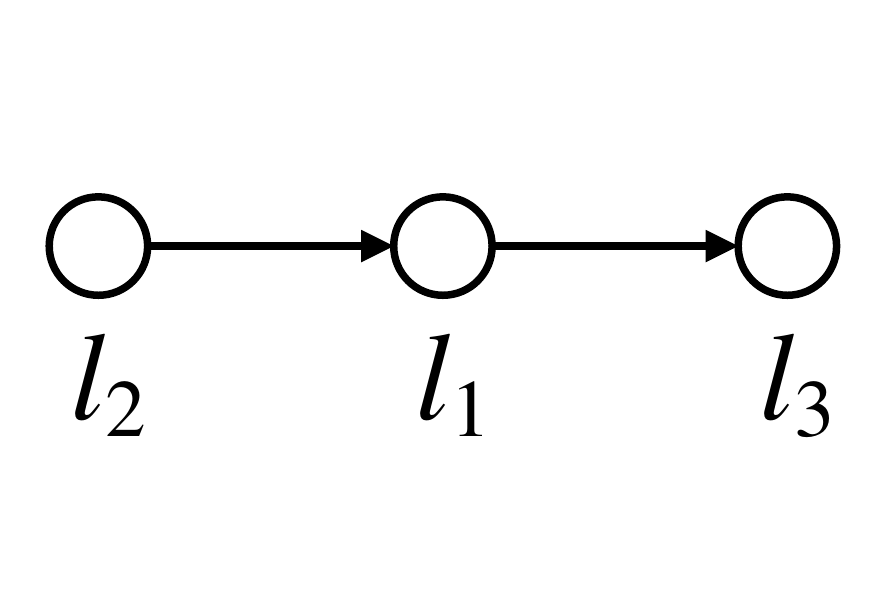}}\quad
\subfloat[BN Structure]{\includegraphics[width=1.3in]{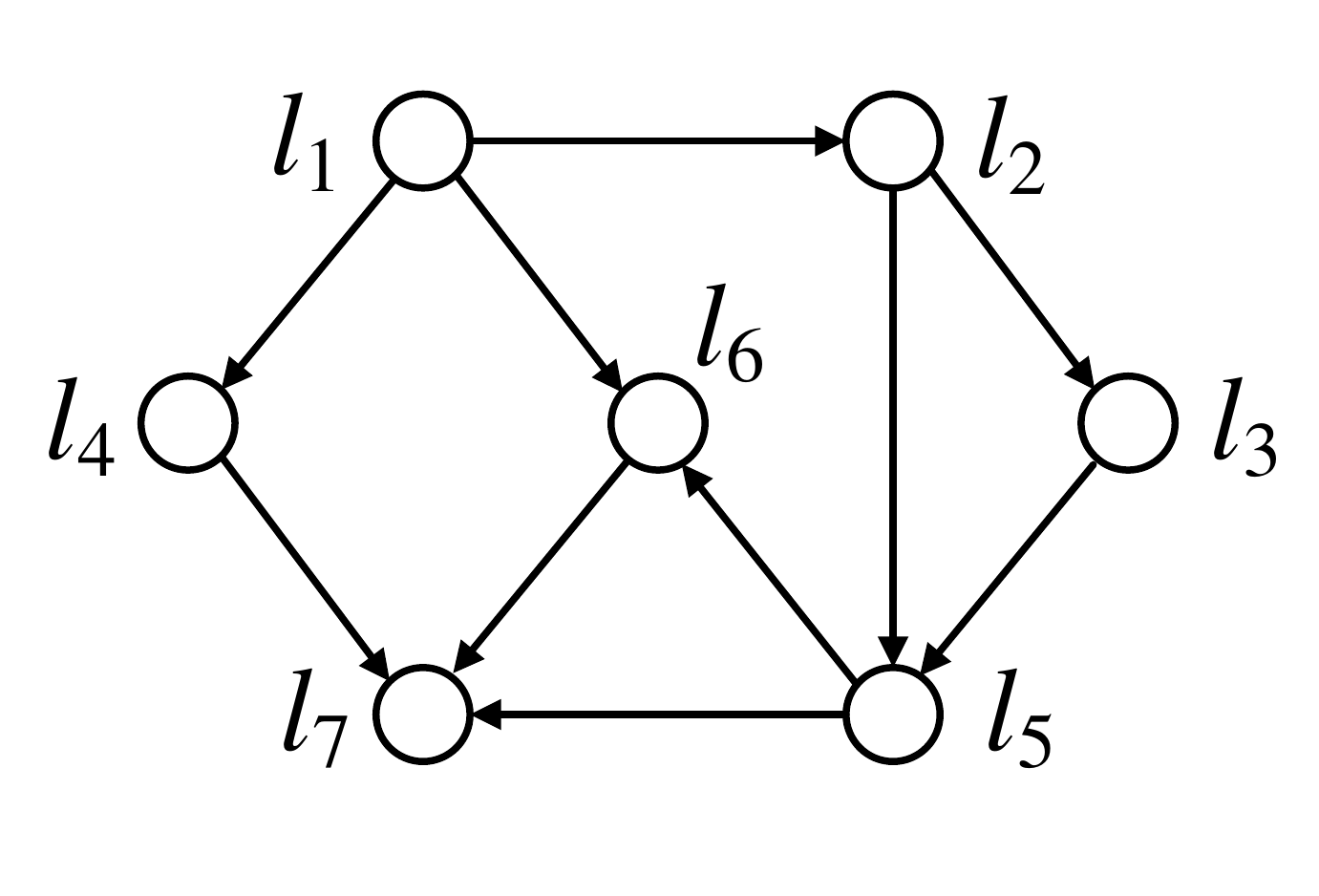}}
\caption{Conditional independence relations in BN.}
\label{fig.IndepBN}
\end{figure}

The conditional independence relation can be proved for the tail-to-tail case and head-to-tail case in BN. As show in Figures~\ref{fig.IndepBN}(a) and~\ref{fig.IndepBN}(b), $l_2$ and $l_3$ are conditionally independent given $l_1$. By applying this rule, the number of parameters in Eq.~(\ref{eq.BN}) can be largely reduced. For example, the joint probability distribution in Figure~\ref{fig.IndepBN}(c) can be simplified as

\begin{equation}\label{eq.BN2}
\centering
\begin{split}
p(y_1,\dots,y_7)=p(y_1)p(y_2|y_1)p(y_3|y_2)\\
\cdots p(y_7|y_4,y_5,y_6).
\end{split}
\end{equation}
As a result, BN defines a unique joint probability distribution over $\mathbb{L}$, which can be generally expressed as

\begin{equation}\label{eq.BN3}
p(y_1,y_2,\dots,y_M)=\prod_{j=1}^M\theta_{y_j|{\rm Pa}(y_j)}=\prod_{j=1}^M p(y_j|{\rm Pa}(y_j)),
\end{equation}
where $\theta$ means a specification of $\Theta$, and ${\rm Pa}(y_j)$ denotes the set of parents of $y_j$ in $\mathbb{G}$. BN can capture the complex correlations among multiple labels, thus, it is a promising technique to improve the performance of CC approach.

\section{The Proposed BNCC Method}\label{sec.proposedModel}

In this section, we will propose the BN model for label correlation analysis and the BNCC method for MLL.

\subsection{Modeling Label Correlations}\label{subsec.labelCorrelation}

\begin{figure}[t]
\centering
\includegraphics[width=2.5in]{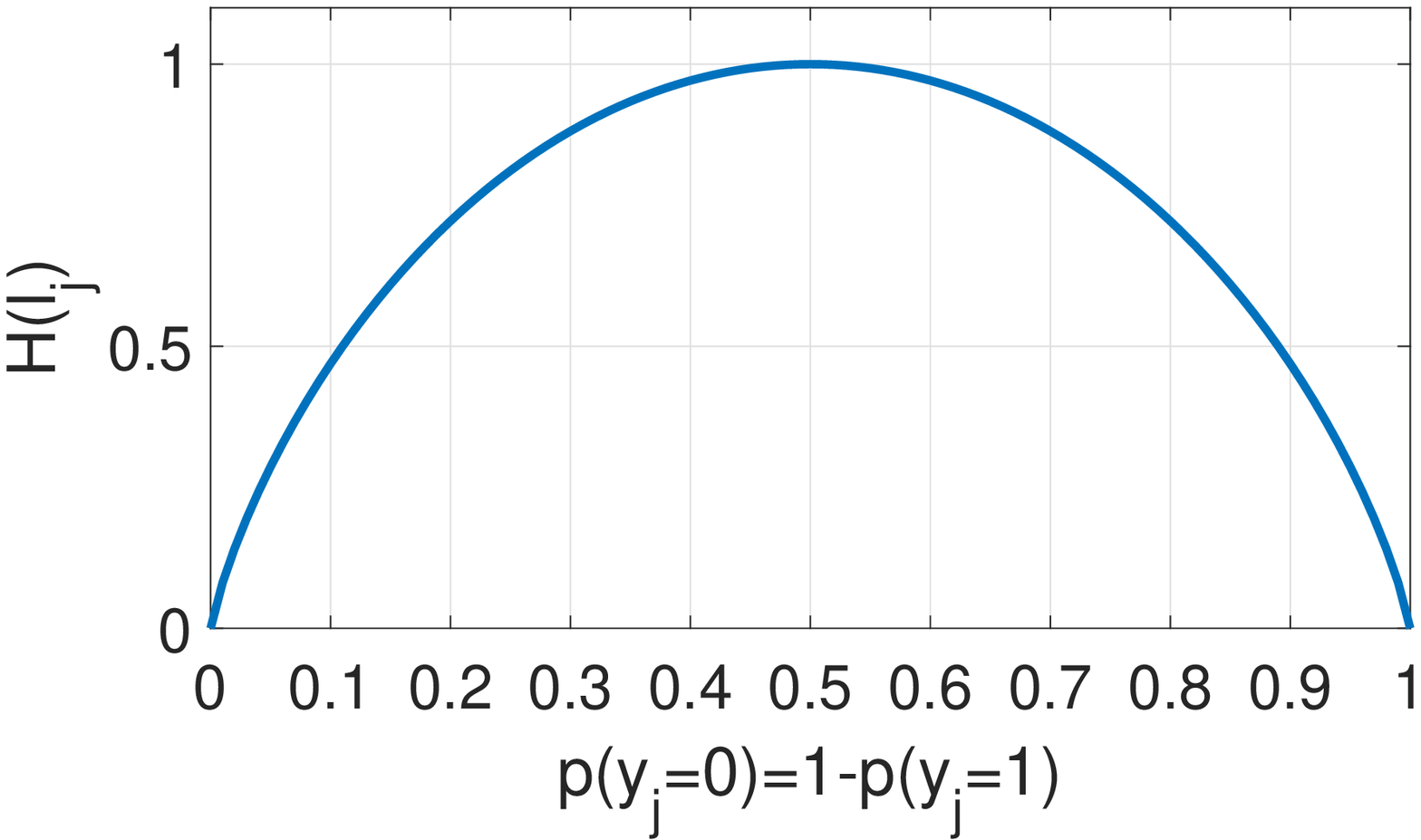}
\caption{Uncertainty of a single label.}
\label{fig.entropy}
\end{figure}

In order to model the correlations among multiple labels, we first give an analysis on the uncertainty of a single label.

\begin{mydef}\label{def.singleU}
(Uncertainty of a Label) Given a single label $l_j\in\mathbb{L}$, the uncertainty of its decision is defined by classical entropy:
\begin{equation}\label{qe.singleU}
\mathcal{H}(l_j)=-\sum_{y_j\in \{0,1\}}p(y_j)\log p(y_j).
\end{equation}
\end{mydef}

As shown in Figure~\ref{fig.entropy}, $\mathcal{H}(l_j)$ will attain its maximum when $l_j$ can be assigned to the positive and negative decisions with an equal probability, and will attain its minimum when $l_j$ can only be assigned to the positive or negative decision. Having this classical definition, we further consider $l_j$ under the condition of another label $l_k$, and raise a question that when the decision of $l_k$ is given, what is the uncertainty of deciding $l_j$? In order to answer this question, the following definition is given.

\begin{mydef}\label{def.conditionalU}
(Conditional Uncertainty of a Label Under Another Label) Given a pair of labels $l_j,l_k\in\mathbb{L}$ ($l_j\neq l_k$), the uncertainty of $l_j$ under the condition of $l_k$, denoted as $\mathcal{H}(l_j|l_k)$, is defined by conditional entropy:
\begin{equation}\label{eq.conditionalU}
\begin{array}{llll}
\mathcal{H}(l_j|l_k)\\
=\sum_{y_k\in\{0,1\}}p(y_k)\mathcal{H}(l_j|y_k)\\
=-\sum_{y_k\in\{0,1\}}p(y_k)\sum_{y_j\in\{0,1\}}p(y_j|y_k)\log p(y_j|y_k)\\
=-\sum_{y_k\in\{0,1\}}\sum_{y_j\in\{0,1\}}p(y_j,y_k)\log p(y_j|y_k)\\
=-\sum_{y_k\in\{0,1\}}\sum_{y_j\in\{0,1\}}p(y_j,y_k)\log\frac{p(y_j,y_k)}{p(y_k)}\\
=\sum_{y_k\in\{0,1\}}\sum_{y_j\in\{0,1\}}p(y_j,y_k)\log\frac{p(y_k)}{p(y_j,y_k)}
\end{array}.
\end{equation}
\end{mydef}

It is easy to prove that the conditional entropy $\mathcal{H}(l_j|l_k)$ in Definition~\ref{def.conditionalU} has the following properties:

\begin{itemize}
\item $\mathcal{H}(l_j|l_k)\in[0,1]$ reflects the amount of information that $l_k$ carries about $l_j$, i.e., the larger value of $\mathcal{H}(l_j|l_k)$, the less information $l_k$ carries about $l_j$, and vice versa;
\item $\mathcal{H}(l_j|l_k)$ attains its minimum $0$ if and only if the value of $l_j$ can be completely predicted by the value of $l_k$;
\item $\mathcal{H}(l_j|l_k)$ attains its maximum $1$ when the value of $l_k$ has no help in deciding the value of $l_j$;
\item $\mathcal{H}(l_j|l_k)\neq \mathcal{H}(l_k|l_j)$, which means that conditional entropy has the property of asymmetry, i.e., the uncertainty of deciding $l_j$ when given $l_k$ is different from the uncertainty of deciding $l_k$ when given $l_j$.
\end{itemize}

Having the above analyses, we know that $\mathcal{H}(l_j|l_k)$ could be treated as the level of difficulty in deciding $l_j$ based on $l_k$, which also reflects the independence degree of $l_j$ on $l_k$.

\begin{mydef}\label{def.directedDepen}
(Dependence Degree) Given a pair of labels $l_j,l_k\in\mathbb{L}$ ($l_j\neq l_k$), the dependence degree of $l_j$ on $l_k$ (also called the influence degree of $l_k$ on $l_j$), denoted by $\mathcal{I}(l_k\rightarrow l_j)$, is defined as:
\begin{equation}\label{eq.directedDepen}
\mathcal{I}(l_k\rightarrow l_j)=1-\mathcal{H}(l_j|l_k).
\end{equation}
\end{mydef}

Definition~\ref{def.directedDepen} provides a way to evaluate the informativeness of a directed edge in a graph, which can help construct an initial structure for BN. Furthermore, both positive relationships, i.e., $p(1,1)$ and $p(0,0)$, and negative relationships, i.e., $p(1,0)$ and $p(0,1)$, are included in Eq.~(\ref{eq.conditionalU}). Thus, it has taken into account all the possible relations between two labels.

The conditional uncertainty of a label can be further extended by revising the condition from a single label $l_k$ to a label subset $\mathbb{L}^*\subset\mathbb{L}$. Then, the following definition is given.

\begin{mydef}\label{def.conditionalU2}
(Conditional Uncertainty of a Label Under a Label Set) Given a single label $l_j\in\mathbb{L}$ and a label subset $\mathbb{L}^*\subset\mathbb{L}$ where $l_j\notin\mathbb{L}^*$, the uncertainty of $l_j$ under the condition of $\mathbb{L}^*$, denoted by $\mathcal{H}(l_j|\mathbb{L}^*)$, is defined as:
\begin{equation}\label{eq.conditionalU2}
\begin{array}{llll}
\mathcal{H}(l_j|\mathbb{L}^*)\\
=\sum_{\mathcal{L}^*\in\{0,1\}^{\mathbb{L}^*}}p(\mathcal{L}^*)\mathcal{H}(l_j|\mathcal{L}^*)\\
=\sum_{\mathcal{L}^*\in\{0,1\}^{\mathbb{L}^*}}\sum_{y_j\in\{0,1\}}p(y_j,\mathcal{L}^*)\log\frac{p(y_k)}{p(y_j,\mathcal{L}^*)}
\end{array},
\end{equation}
where $\mathcal{L}^*$ represents a possible value configuration for the labels in $\mathbb{L}^*$.
\end{mydef}

Similarly, the dependence degree of $l_j$ on $\mathbb{L}^*$ (also called the influence degree of $\mathbb{L}^*$ on $l_j$), denoted by $\mathcal{I}(\mathbb{L}^*\rightarrow l_j)$, is defined as:

\begin{equation}\label{eq.directedDepen2}
\mathcal{I}(\mathbb{L}^*\rightarrow l_j)=1-\mathcal{H}(l_j|\mathbb{L}^*).
\end{equation}

\subsection{Designing Scoring Function for BN}\label{subsec.score}

Having the definitions mentioned above, a fully connected DCG can be build up by linking each pair of labels in two-way, denoted as $\mathbb{G}^{\circ}=(\mathbb{V},\mathbb{E}^{\circ})$. There will be two links between every pair of labels $l_j$ and $l_k$, i.e., $l_j\rightarrow l_k$ and $l_k\rightarrow l_j$. We know that $\mathcal{I}(l_k\rightarrow l_j)$ is the weight of link $l_k\rightarrow l_j$, which represents the dependence degree of $l_j$ on $l_k$, and $\mathcal{I}(l_j\rightarrow l_k)$ is the weight of link $l_j\rightarrow l_k$, which represents the dependence degree of $l_k$ on $l_j$. Thus, there will be $2\times\mathbf{C}_{M}^{2}$ edges with many directed cycles in $\mathbb{G}^{\circ}$, which is a complex structure to be optimized. The following two issues need to be resolved:
\begin{enumerate}
\item given a DCG, how to simplify it into an effective BN structure without cycles;
\item how to define a scoring function to evaluate a BN structure.
\end{enumerate}
In order to handle the first issue, we refine a DCG into a DAG by breaking cycles iteratively. As we know, larger value of $\mathcal{I}(l_k\rightarrow l_j)$ represents stronger dependence of $l_j$ on $l_k$. Thus, we iteratively break a cycle through removing the edge with the lowest dependence degree in the cycle, until a DAG is obtained. The detailed steps are described in Algorithm~\ref{alg.removeCycles}.

\begin{algorithm}[t]
\caption{Refine a DCG into a DAG}
\label{alg.removeCycles}
\KwIn{Original DCG, denoted as $\mathbb{G}^{\circ}=(\mathbb{V},\mathbb{E}^{\circ})$;\\
\quad\quad\quad\ \ Dependence degree between any pair of labels, denoted as $\mathcal{I}$ .}
\KwOut{Refined DAG, i.e., $\mathbb{G}=(\mathbb{V},\mathbb{E})$.}

Let $\mathbb{G}=\mathbb{G}^{\circ}$;\\
\While{$\mathbb{G}$ has circles}{
Find a circle in $\mathbb{G}$, denoted as $Cyc$;\\
Let $\mathbb{E}^*$ contain all the edges in $Cyc$, i.e., $\mathbb{E}^*=\{e|e\in Cyc\}$;\\
Find the edge in $\mathbb{E}^*$ with minimum dependence degree $e^*=\argmin_{e\in \mathbb{E}^*}\mathcal{I}(e_{\rm start}\rightarrow e_{\rm end})$;\\
Delete $e^*$ from $\mathbb{G}$;
}

\Return{$\mathbb{G}$}.\\
\end{algorithm}

As for the second issue, the most well-known scoring function for BN is the Bayesian information criterion (BIC), i.e.,

\begin{equation}\label{eq.score}
S_{\mathbb{D}}(\mathbb{G})=L_\mathbb{D}(\mathbb{G})-\frac{{\rm Dim}_\mathbb{G}}{2}\log N.
\end{equation}

In Eq.~(\ref{eq.score}), the first term is the log-likelihood of graph $\mathbb{G}$ with respect to the training set $\mathbb{D}$, which reflects the fitting degree between $\mathbb{G}$ and the data. In this paper, since the purpose is to model the correlations among multiple labels through BN, we re-define $L_\mathbb{D}(\mathbb{G})$ by using the dependence degree proposed in section~\ref{subsec.labelCorrelation}. Finally, $L_{\mathbb{D}}(\mathbb{G})$ can be written as the sum of the scores for each label $l_j$, i.e.,

\begin{equation}\label{eq.scoreCor}
\begin{array}{llll}
L_\mathbb{D}(\mathbb{G})\\
=N\sum\limits_{j=1}^{M}(\mathcal{I}(\mathbb{L}^*\rightarrow l_j))\\
=N\sum\limits_{j=1}^{M}(1-\mathcal{H}(l_j|{\rm Pa}(l_j)))\\
=N(1-\sum\limits_{j=1}^{M}\sum\limits_{q=1}^{Q_j}p({\rm Pa}_q(l_j))\mathcal{H}(l_j|{\rm Pa}_q(l_j)))\\
=N+N\sum\limits_{j=1}^{M}\sum\limits_{q=1}^{Q_j}\sum\limits_{y_j\in\{0,1\}}p(y_j,{\rm Pa}_q(l_j))\log\frac{p(y_j,{\rm Pa}_q(y_j))}{p({\rm Pa}_q(y_j))}
\end{array},
\end{equation}
where $Q_j$ represents the number of possible value configurations for ${\rm Pa}(l_j)$, and ${\rm Pa}_q(l_j)$ represents the $q$-th configuration. Furthermore, given data set $\mathbb{D}$, the probability can be calculated by frequency, thus we have

\begin{equation}\label{eq.scoreCor2}
\begin{array}{llll}
L_\mathbb{D}(\mathbb{G})\\
=N+N\sum\limits_{j=1}^{M}\sum\limits_{q=1}^{Q_j}\sum\limits_{y_j\in\{0,1\}}\frac{N_{jq}^{(y_j)}}{N}\log\frac{N_{jq}^{(y_j)}}{N_{jq}}\\
=N+\sum\limits_{j=1}^{M}\sum\limits_{q=1}^{Q_j}\sum\limits_{y_j\in\{0,1\}}N_{jq}^{(y_j)}\log\frac{N_{jq}^{(y_j)}}{N_{jq}}
\end{array},
\end{equation}
where $N_{jq}$ indicates the number of instances in $\mathbb{D}$ with the $q$-{th} configuration for ${\rm Pa}(l_j)$, and $N_{jq}^{(y_j)}$ indicates the number of instances in $\mathbb{D}$ with $l_j=y_j$ given the $q$-th configuration for ${\rm Pa}(l_j)$.

In Eq.~(\ref{eq.score}), the second term is a complexity penalization factor, which prevents over-fitting between $\mathbb{G}$ and the training data. It can be denoted by the number of independent parameters in the structure, i.e., ${\rm Dim}_\mathbb{G}=\sum_{j=1}^{M}Q_j$. Finally, the scoring function becomes
\begin{equation}\label{eq.score2}
\begin{array}{llll}
S_{\mathbb{D}}(\mathbb{G})\\
=N+\sum\limits_{j=1}^{M}\sum\limits_{q=1}^{Q_j}\sum\limits_{y_j\in\{0,1\}}N_{jq}^{(y_j)}\log\frac{N_{jq}^{(y_j)}}{N_{jq}}-\sum\limits_{j=1}^{M}\frac{Q_j}{2}\log N.
\end{array}
\end{equation}

\subsection{Learning the Optimal BN Structure}

In order to learn the optimal BN based on a given data set, all possible structures can be considered as a domain, and the scoring function is used to measure the quality of a structure in the domain. Finding the best BN structure is equivalent to find the maximum value of the scoring function. However, the number of candidate structures is huge, it is impossible to traverse all of them. Since the scoring function can be decomposed with regard to different labels $l_j$ ($j=1,\ldots,M$), we can find the optimal parent set for each individual label separately. In this case, an initial label order should be determined, such that the labels can be optimized one-by-one. Topological sorting~\cite{khan1962topological} gives a good solution for ordering the nodes in a DAG, which is presented in Algorithm~\ref{alg.nodesSorting}.

\begin{algorithm}[t]
\caption{Topological Sorting for Nodes in a DAG}
\label{alg.nodesSorting}
\KwIn{DAG $\mathbb{G}=(\mathbb{V},\mathbb{E})$, where $\mathbb{V}=\{l_1,l_2,\ldots,l_M\}$.}
\KwOut{Label order $\alpha:{l_{\alpha(1)},l_{\alpha(2)},\ldots,l_{\alpha(M)}}$.}

Let $\mathbb{V}^*=\mathbb{V}$;\\
\For{each node $l$ in $\mathbb{V}^*$}{
Calculate the in-degree of $l$ in $\mathbb{G}$, denoted as $\mathfrak{in}(l)$;\\
}
\For{$j=1$ to $M$}{
Select a node from $\mathbb{V}^*$ with $\mathfrak{in}(l)=0$, denoted as $l^*$;\\
Let $l_{\tau(j)}=l^*$, $\mathbb{V}^*=\mathbb{V}^*\setminus l^*$;\\
\For{each node $l$ pointed to by $l^*$}{
Let $\mathfrak{in}(l)=\mathfrak{in}(l)-1$;\\
}}

\Return{Label order $\alpha:{l_{\alpha(1)},l_{\alpha(2)},\dots,l_{\alpha(M)}}$}.\\
\end{algorithm}

Having the label order derived from Algorithm~\ref{alg.nodesSorting}, the optimal parent set can be discovered for each label separately. In~\cite{cooper1992bayesian} and~\cite{bouckaert1994probabilistic}, two algorithms called {\it K2} and {\it K3} were developed for this purpose, they begin with an empty parent set for each label and add new members to the set based on a greedy strategy. In this paper, we propose a similar strategy by applying the newly proposed scoring function.

\begin{figure}[t]
\centering
\subfloat[Structure 1]{\includegraphics[height=1.5in]{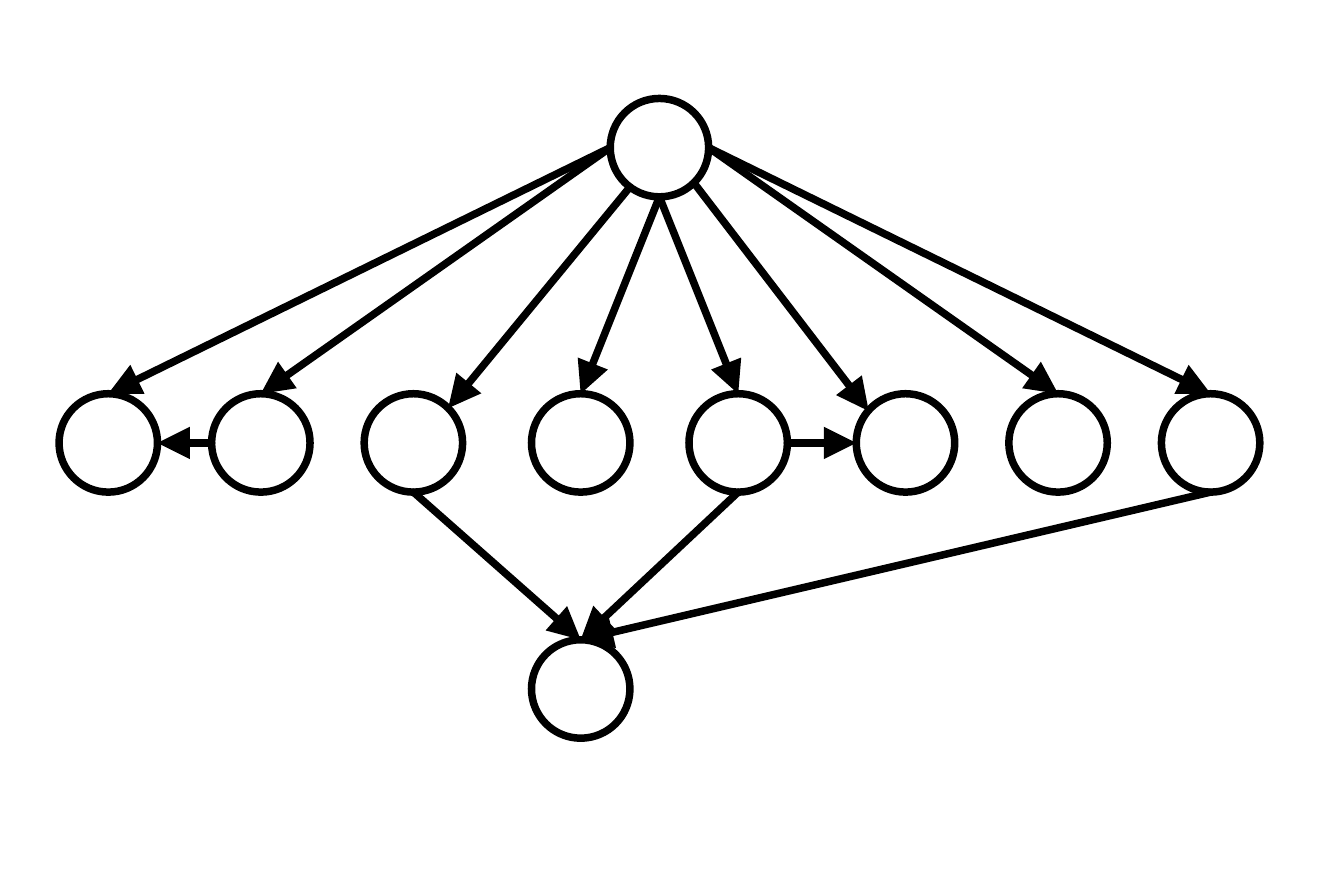}}
\subfloat[Structure 2]{\includegraphics[height=1.5in]{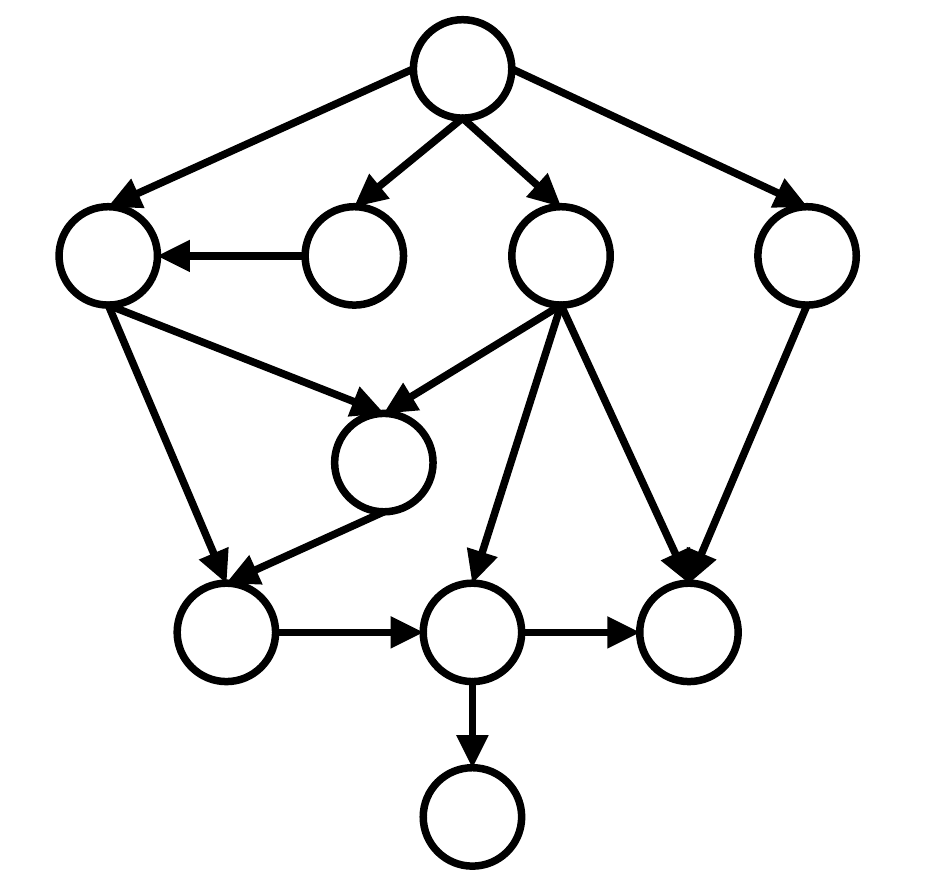}}
\caption{Two BN structures for a ten-label problem.}
\label{fig.twoBNs}
\end{figure}

Due to the decomposition characteristic of the scoring function, the optimal graph can be constructed by building and merging the optimal sub-graphs for individual labels. Let $\mathbb{G}_{l|{\rm Pa}(l)}$ be the sub-graph that is only composed of $l$ and ${\rm Pa}(l)$, and assume that an initial label order $\alpha:$ $l_{\alpha(1)}$, $l_{\alpha(2)}$, $\dots$, $l_{\alpha(M)}$ is produced by Algorithm~\ref{alg.nodesSorting}. Then, the details for optimal parent set determination are described in Algorithm~\ref{alg.learningBN}. Several key points are highlighted as follows.
\begin{itemize}
\item For $l_{\alpha(j)}$, $j=1,2,\ldots,M$, the parent set ${\rm Pa}(l_{\alpha(j)})$ is initialized as $\emptyset$ and the candidate set ${\rm Pred}_j$ is initialized as $\mathbb{L}\setminus l_{\alpha(j)}$. The candidate in ${\rm Pred}_j$ that can maximize the gain of score for $\mathbb{G}_{l_{\alpha(j)}|{\rm Pa}(l_{\alpha(j)})}$ will be added to ${\rm Pa}(l_{\alpha(j)})$ iteratively.
\item Structural anomalies may appear in the learned graph, i.e., one node may have too many child nodes, causing the structure to be over-wide and obstructing the graph from including more meaningful edges. As shown in Figure~\ref{fig.twoBNs}, structure 2 is more ideal than structure 1 since structure 2 captures more diverse relationships. In this case, we control the number of child nodes by setting a threshold.
\item It is known from literature that in the optimal BN, each label has at most $\log N$ parents. Thus, we terminate the selection process for $l_{\alpha(j)}$ when $|{\rm Pa}(l_{\alpha(j)})|> \log N$.
\end{itemize}

\begin{algorithm}[h]
\footnotesize
\caption{Learning the Optimal Parent Set For Each Node}
\label{alg.learningBN}
\KwIn{Training set $\mathbb{D}=\{(\mathbf{x}_i,\mathbf{y}_i)\}_{i=1}^{N}$ and label set $\mathbb{L}=\{l_1,l_2,\ldots,l_M\}$;\\
\quad\quad\quad\ \ Initial label order $\alpha:{l_{\alpha(1)},l_{\alpha(2)},\ldots,l_{\alpha(M)}}$;\\
\quad\quad\quad\ \ Threshold $n$ for maximum number of child nodes.}
\KwOut{Graph $\mathbb{G}_{\rm new}^{\circ}$ with optimized parent label sets.}

Let $\mathbb{G}_{\rm new}^{\circ}$ be an empty graph;\\
\For{$j=1$ to $M$}{
Let ${\rm Pred}_j=\mathbb{L}\setminus l_{\alpha(j)}$;\\
Let ${\rm Pa}(l_{\alpha(j)})=\emptyset$;\\
\For{each $l\in{\rm Pred}_j$}{
\If{$l$ already has $n$ child nodes}{
${\rm Pred}_j={\rm Pred}_j\setminus l$;\\
}}
Let $S_{\rm opt}=-\infty$;\\
\While{$|{\rm Pa}(l_{\alpha(j)})|\leq \log N$ and ${\rm Pred}_j\neq\emptyset$}{
Let $l^*=\argmax_{l\in{\rm Pred}_j} S_{\mathbb{D}}(\mathbb{G}_{l_{\alpha(j)}|{\rm Pa}(l_{\alpha(j)})\bigcup l})$;\\
Let $S_{\rm new}=S_{\mathbb{D}}(\mathbb{G}_{l_{\alpha(j)}|{\rm Pa}(l_{\alpha(j)})\bigcup l^*})$;\\
\uIf{$S_{\rm new} > S_{\rm opt}$}{
${\rm Pa}(l_{\alpha(j)})={\rm Pa}(l_{\alpha(j)}) \bigcup {{l^*}}$;\\
${\rm Pred}_j={\rm Pred}_j\setminus l^*$;\\
$S_{\rm opt}=S_{\rm new}$;
}\Else{
{\rm\bf break};
}}
Update $\mathbb{G}_{\rm new}^{\circ}$ by merging $\mathbb{G}_{l_{\alpha(j)}|{\rm Pa}(l_{\alpha(j)})}$ into $\mathbb{G}_{\rm new}^{\circ}$;
}

\Return{$\mathbb{G}_{\rm new}^{\circ}$}.
\end{algorithm}

It is noteworthy that the graph produced by Algorithm~\ref{alg.learningBN} may have cycles. Since the value of dependence degree is non-negative, and the larger the dependence degree of a label on its parents, the larger the score will be, i.e.,

\begin{equation}
\mathcal{I}(l_k\rightarrow l_j) > \mathcal{I}(l_q\rightarrow l_j)\Rightarrow S(\mathbb{G}_{l_j|l_k}) > S(\mathbb{G}_{l_j|l_q}),
\end{equation}
in this case, Algorithm~\ref{alg.removeCycles} can be used again to refine the newly constructed DCG into an effective BN, and the final label order is obtained by applying Algorithm~\ref{alg.nodesSorting} on the BN.

\subsection{Constructing CC Model}

In this section, we summarize the process of building the CC model by learning the optimal BN. First, the dependence degree between any pair of labels is computed based on Definition~\ref{def.directedDepen}, and a fully connected DCG $\mathbb{G}^{\circ}$ is constructed. Then, by breaking cycles in $\mathbb{G}^{\circ}$, an initial DAG $\mathbb{G}$ is produced and an initial label order is obtained by applying topological sorting on the nodes of $\mathbb{G}$. The optimal parent set for each label is learned following the initial order, and the learned sub-graphs for different labels are merged to be a new DCG $\mathbb{G}^{\circ}_{\rm new}$. Afterwards, the same process for breaking cycles is performed on $\mathbb{G}^{\circ}_{\rm new}$, and a new DAG $\mathbb{G}_{\rm new}$ is produced. The final label order is obtained by applying topological sorting on the nodes of $\mathbb{G}_{\rm new}$. The entire procedure for building the CC model is summarized in Algorithm~\ref{alg.buildingCC}.

\begin{algorithm}[h]
\caption{Constructing the BN-based CC Model}
\footnotesize
\label{alg.buildingCC}
\KwIn{Training set $\mathbb{D}=\{(\mathbf{x}_i,\mathbf{y}_i)\}_{i=1}^{N}$ and label set $\mathbb{L}=\{l_1,l_2,\ldots,l_M\}$;\\
\quad\quad\quad\ \ Threshold $n$ for maximum number of child nodes.}
\KwOut{CC model $h_1,h_2,\ldots,h_M$.}

Calculate the dependence degree in two-way between any pair of labels based on Eq.~(\ref{eq.directedDepen}), denoted as $\mathcal{I}$;\\
Build up a fully connected DCG with all labels, i.e., $\mathbb{G}^{\circ}$;\\
Call Algorithm~\ref{alg.removeCycles} on $\mathbb{G}^{\circ}$ and $\mathcal{I}$ to get initial DAG $\mathbb{G}$;\\
Call Algorithm~\ref{alg.nodesSorting} on $\mathbb{G}$ to get the initial label order $\alpha:\ l_{\alpha(1)},l_{\alpha(2)},\ldots,l_{\alpha(M)}$;\\
Call Algorithm~\ref{alg.learningBN} on $\mathbb{D}$, $\alpha$ and $n$ to learn the new graph $\mathbb{G}^{\circ}_{\rm new}$ with optimized parent label set for each label;\\
Call Algorithm~\ref{alg.removeCycles} on $\mathbb{G}^{\circ}_{\rm new}$ and $\mathcal{I}$ to get optimized DAG $\mathbb{G}_{\rm new}$;\\
Call Algorithm~\ref{alg.nodesSorting} on $\mathbb{G}_{\rm new}$ to get the final label order $\tau:\ l_{\tau(1)},l_{\tau(2)},\ldots,l_{\tau(M)}$;\\
Train the CC model $h_1,h_2,\ldots,h_M$ based on $\mathbb{D}$ and $\tau$;\\

\Return CC model $h_1,h_2,\ldots,h_M$.\\

\end{algorithm}

\subsection{Complexity Analysis}

It is easy to analyze that in Algorithm~\ref{alg.removeCycles}, breaking a cycle $Cyc$ has the linear complexity of $O(|Cyc|)$, where $|Cyc|$ is the number of edges in $Cyc$, and in Algorithm~\ref{alg.nodesSorting}, the topological sorting on $M$ nodes has the maximum complexity of $O(M^2)$. As for Algorithm~\ref{alg.learningBN}, the optimal parent set determination process is performed regarding each label separately. For each label, the main time cost focuses on the while loop, i.e., steps 11$\sim$21. It is noteworthy that computing the scoring function, i.e., Equation~(\ref{eq.score2}), has the complexity of $O(MQ+M)$, where $M$ is the number of labels and $Q$ is the number of possible value configurations for the parents of a label. Let $u=MQ+M$ be a unit time cost, thus, during each while loop, computing the scoring function for all candidate parent nodes of $l_{\alpha(j)}$ has the complexity of $O(|{\rm Pred}_j|u)$, and selecting the best candidate has the complexity of $O(|{\rm Pred}_j|)$, where ${\rm Pred}_j$ contains all the remained candidates in the current iteration. The initial size of ${\rm Pred}_j$ is $M-1$ and it is reduced by one during each iteration. Since there are $\log N$ iterations at most, the while loop for each label has the maximum complexity of $O(Mu+(M-1)u+\ldots+(M-\log N+1)u)$.

\section{Empirical Studies}\label{sec.experiment}

In this section, we conduct extensive experiments to show the effectiveness and efficiency of the proposed method.

\subsection{Methods and Data Sets for Performance Comparison}

Five state-of-the-art MLL methods, as well as the proposed approach, are listed in this section for performance comparison.

\paragraph{Binary Relevance ({\sf BR})} {\sf BR} is a baseline method that decomposes the multi-label problem into multiple independent binary problems and each binary problem corresponds to a single label~\cite{tao2007multilabel}.

\paragraph{Calibrated Label Ranking ({\sf CLR})} {\sf CLR} transforms the multi-label problem into a label ranking problem by building a binary classifier for each pair of labels~\cite{Furnkranz2008multilabel}.

\paragraph{Classifier Chain ({\sf CC})} Traditional {\sf CC} approach transforms the multi-label problem into a chain of binary problems based on a randomly generated label order~\cite{read2009classifier}.

\paragraph{Group sensitive Classifier Chain ({\sf GCC})} The basic idea of {\sf GCC} is to cluster the training instances into $m$ different groups and learn the label correlation locally for each group to build a classifier chain~\cite{huang2015group}. The best parameter reported in the literature $m=5$ is adopted.

\paragraph{Ensemble Classifier Chains ({\sf ECC})} This method trains multiple CCs based on different random orders, and gets the final prediction by collective voting~\cite{read2011classifier}. We set the number of CCs as 10 for the smaller data sets and set the number of CCs as 5 for the larger data sets.

\paragraph{Bayesian Network based Classifier Chain ({\sf BNCC})} The proposed approach is realized.

Eighteen data sets are selected for performance comparison. These data sets are collected from multiple sources including Mulan\footnote{http://mulan.sourceforge.net/datasets.html},
Lamda\footnote{http://lamda.nju.edu.cn/Data.ashx\#data}, and Yahoo\footnote{http://www.kecl.ntt.co.jp/as/members/ueda/yahoo}, which cover a wide range of application domains such as image recognition, music retrieval, text classification, and bio-informatics, etc. The details of the selected data sets are listed in Table~\ref{tab.datasets}.

\begin{table}[!htbp]
\caption{Data Sets for Performance Comparison}
\begin{center}
\scalebox{0.7}{
\begin{tabular}{lccccccc}
\hline
\textbf{Data Sets} &{\it Domain} &{\it \# Training} &{\it \# Testing} &{\it \# Attributes} &{\it \# Labels}  &{\it Card} &{\it AvgIR}\\
\hline
Bibtex	    &Text	      &4,880  &2,515	&1,836	&159	&2.402     &12.498\\
CHD49       &Medicine     &555    &---	    &49     &6      &2.580	   &5.766\\
Emotions	&Music	      &391    &202	    &72	    &6	    &1.869     &1.478\\
Enron	    &Text	      &1,123  &579	    &1,001	&53     &3.378     &73.953\\
EukaryoteGo &Biology      &4,658  &3,108    &12,689 &22     &1.146     &45.012\\
Flags	    &Image        &129    &65	    &19	    &7      &3.392     &2.255\\
Genbase	    &Biology	  &463    &199	    &1,186	&27     &1.252     &37.315\\
GramNegative    &Biology      &836    &556      &1,717  &8      &1.046     &18.448\\
GramPositive    &Biology      &311    &208      &912    &4      &1.008     &3.861\\
HumanGoB3106    &Biology      &3,106    &---	&9,844  &14     &1.185     &15.289\\
HumanPseACC3106 &Biology      &1,862    &---	&440    &14     &1.185     &15.289\\
Mediamill	&Video	      &30,993 &12,914	&120	&101	&4.376     &256.405\\
Medical	    &Text	      &333    &645	    &1,449	&45	    &1.245     &89.501\\
PlantGoB978     &Biology      &588    &390      &3,091  &12     &1.079     &6.690\\
Scene	    &Image	      &1,211  &1,196	&294	&6	    &1.074     &1.254\\
VirusGo     &Biology      &124    &83       &749    &6      &1.217     &4.041\\
WaterQuality&Chemistry    &1,060  &---	    &16     &14     &5.073     &1.767\\
Yeast	    &Biology	  &1,500  &917	    &103	&14	    &4.237     &7.197\\
\hline\noalign{\smallskip}
\end{tabular}}
\begin{minipage}{13.5cm}
\scriptsize{\textbf{Note:} {\it Card} represents label cardinality, i.e., the average number of positive labels associated with the samples; and {\it AvgIR} measures the average degree of imbalance of all labels, the greater avgIR, the greater the imbalance of the data set.}
\end{minipage}
\label{tab.datasets}
\end{center}
\vspace{-0.5cm}
\end{table}

\subsection{Evaluation Metrics}

As observed from the last two columns of Table~\ref{tab.datasets}, multi-label data sets are usually imbalanced, i.e., the number of negative instances is much larger than the number of positive instances for each label. Usually, the purpose of MLL is to correctly identify the positive labels for unseen instances, thus, traditional measures like testing accuracy might be ineffective. In this paper, we utilize four metrics as follows.

\subsubsection{Hamming Loss (${\rm HammingLoss}$)} ${\rm HammingLoss}$ is an instance-based metric that evaluates how many times the instance-label pairs are misclassified, i.e.,
\begin{equation}\label{eq.hammingLoss}
{\rm HammingLoss}=\frac{1}{N}\sum_{i=1}^{N}\frac{1}{M}\sum_{j=1}^{M}\llbracket\mathbf{y}_{ij}\neq\hat{\mathbf{y}}_{ij}\rrbracket,
\end{equation}
where $\llbracket\cdot\rrbracket$ returns 1 or 0 if the internal condition holds or not, $M$ is the number of labels, $N$ is the number of instances, $\mathbf{y}_i$ and $\hat{\mathbf{y}}_i$ are the ground-truth label vector and the predicted label vector for the $i$-th instance. ${\rm HammingLoss}$ treats the positive and negative labels as equally important. As mentioned above, due to the imbalance nature, it is highly biased by the results on the negative labels, thus is not a good measure for MLL.

\subsubsection{Instance-based F-score (${\rm Fscore}$)} ${\rm Fscore}$ is also an instance-based metric that computes a harmonic mean between precision and recall for each instance, i.e.,
\begin{equation}\label{eq.Fscore}
{\rm Fscore} = \frac{1}{N} \sum_{i=1}^{N}\frac{2{\rm Precision}_i\cdot{\rm Recall}_i}{{\rm Precision}_i+{\rm Recall}_i},
\end{equation}
where ${\rm Precision}_i=$ ${\sum_{j=1}^{M}\llbracket\mathbf{y}_{ij}=1\&\&\hat{\mathbf{y}}_{ij}=1\rrbracket} / {\sum_{j=1}^{M}\llbracket\mathbf{y}_{ij}=1\rrbracket}$ \normalsize and ${\rm Recall}_i=$ ${\sum_{j=1}^{M}\llbracket\mathbf{y}_{ij}=1\&\&\hat{\mathbf{y}}_{ij}=1\rrbracket} / {\sum_{j=1}^{M}\llbracket\hat{\mathbf{y}}_{ij}=1\rrbracket}$ are respectively the ratios of correctly predicted positive labels in the real positive set and in the set that has been predicted as positive for the $i$-th instance. Since this metric focuses on the correctly predicted positive labels for each instance, it is much more rational than ${\rm HammingLoss}$.

\subsubsection{Label-based Macro F-score (${\rm MacF}$)} ${\rm MacF}$ is a label-based metric that evaluates the macro average of precision and recall regarding different labels, i.e.,
\begin{equation}\label{eq.macroF}
{\rm MacF} = \frac{1}{M}\sum_{j=1}^{M}\frac{2{\rm TP}_{j}}{2{\rm TP}_{j}+{\rm FP}_{j}+{\rm FN}_{j}},
\end{equation}
where ${\rm TP}_j$, ${\rm FP}_j$, and ${\rm FN}_j$ are the numbers of true positives, false positives, and false negatives regarding the $j$-th label. It focuses on the correct positive predictions regarding each label, thus it is a rational measure for MLL.

\subsubsection{Label-based Micro F-score (${\rm MicF}$)} ${\rm MicF}$ is also a label-based metric, which is a revised version of ${\rm MacF}$, i.e.,
\begin{equation}\label{eq.microF}
{\rm MicF} = \frac{2\sum_{j=1}^{M}{\rm TP}_{j}}{2\sum_{j=1}^{M}{\rm TP}_{j}+\sum_{j=1}^{M}{\rm FP}_{j}+\sum_{j=1}^{M}{\rm FN}_{j}},
\end{equation}
where ${\rm TP}_j$, ${\rm FP}_j$, and ${\rm FN}_j$ are same as in Eq.~(\ref{eq.macroF}). ${\rm MicF}$ also focuses on the correct positive predictions, thus it is a rational measure for MLL.

Obviously, smaller value of ${\rm HammingLoss}$ and larger values of the three F-scores represent better performance of the model.

\subsection{Experimental Setup}

\begin{figure*}[t]
\centering
\subfloat[${\rm HammingLoss}$]{\includegraphics[width=2.5in]{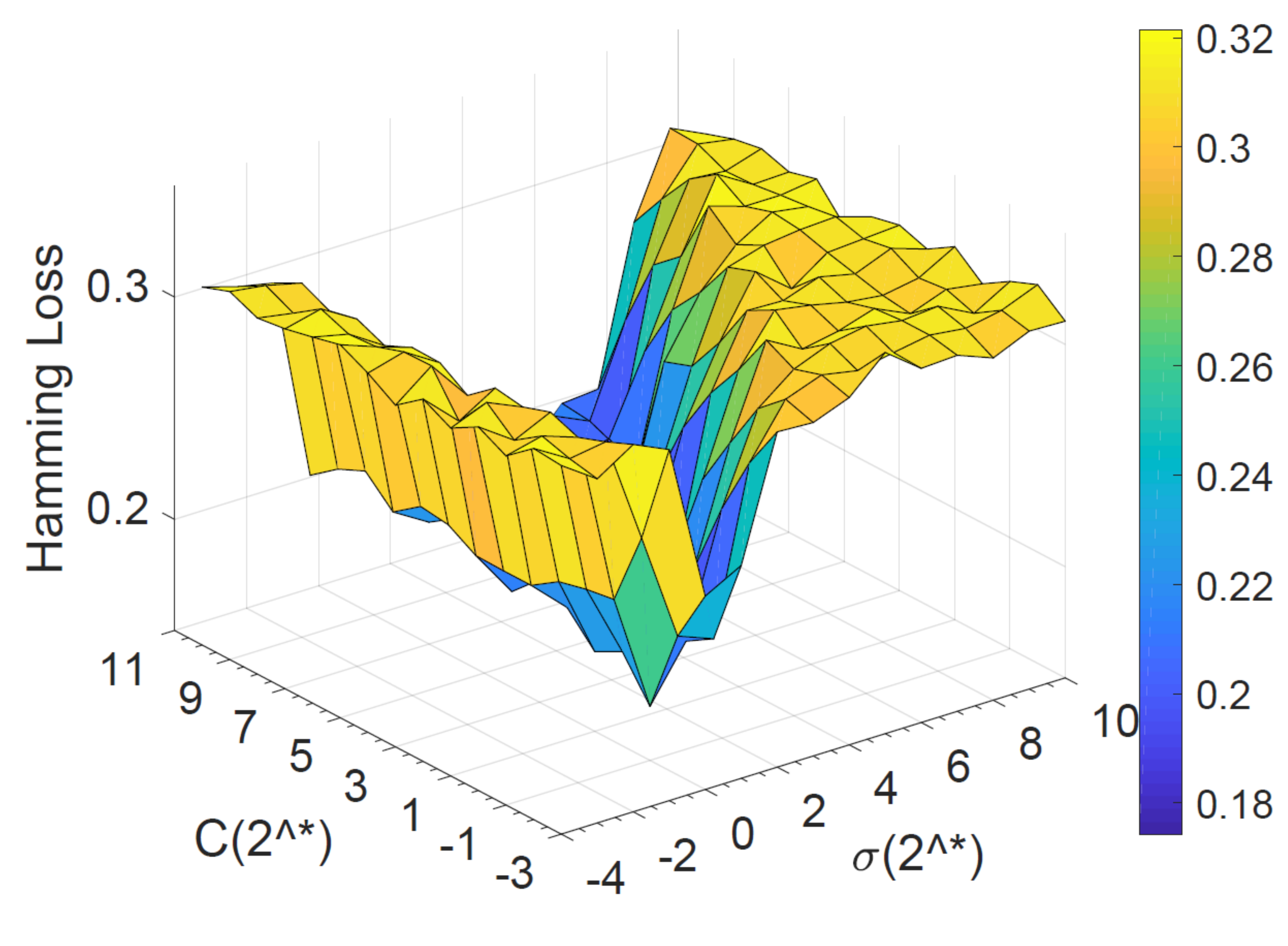}}\quad
\subfloat[${\rm Fscore}$]{\includegraphics[width=2.5in]{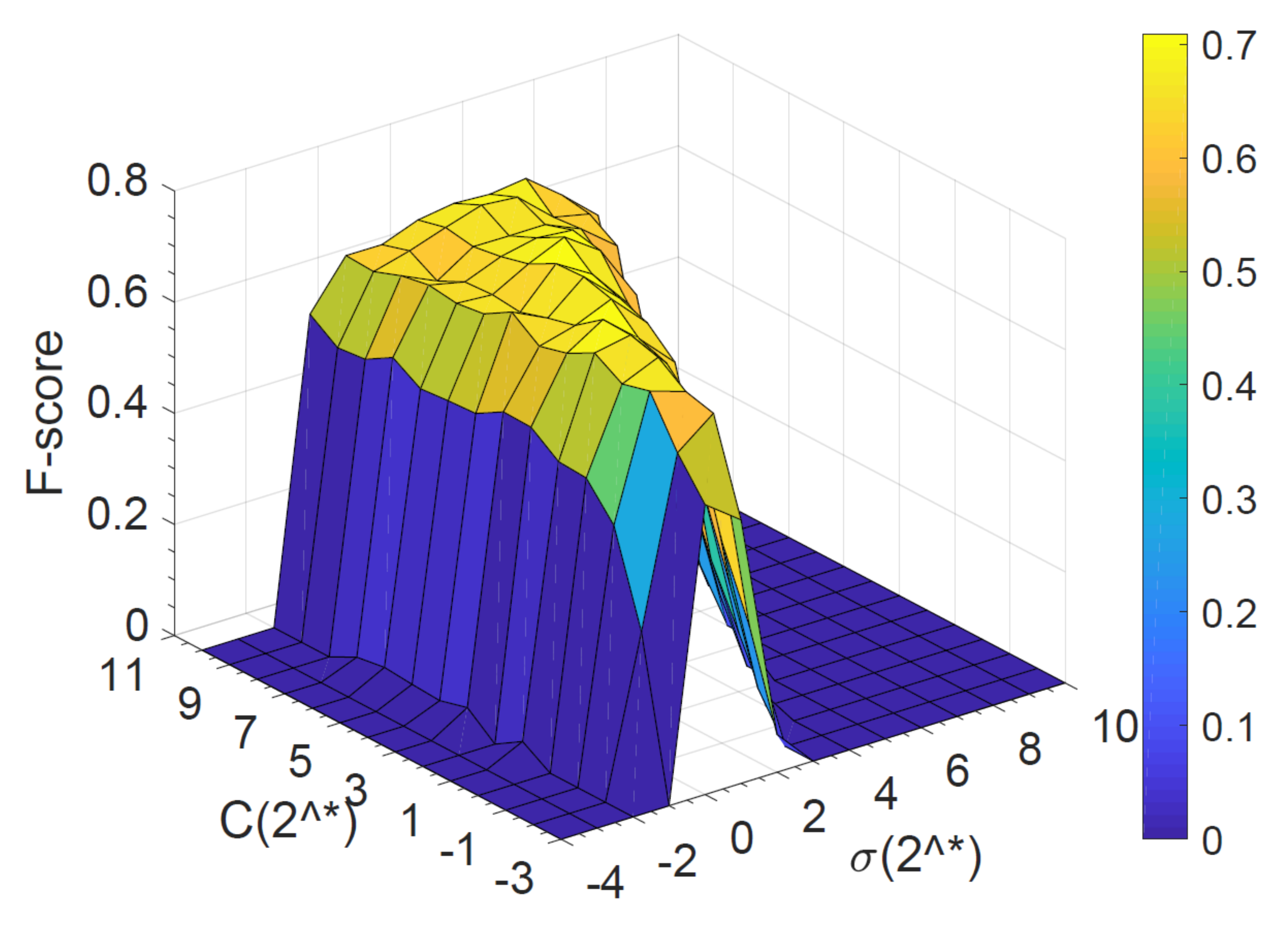}}
\subfloat[${\rm MacF}$]{\includegraphics[width=2.5in]{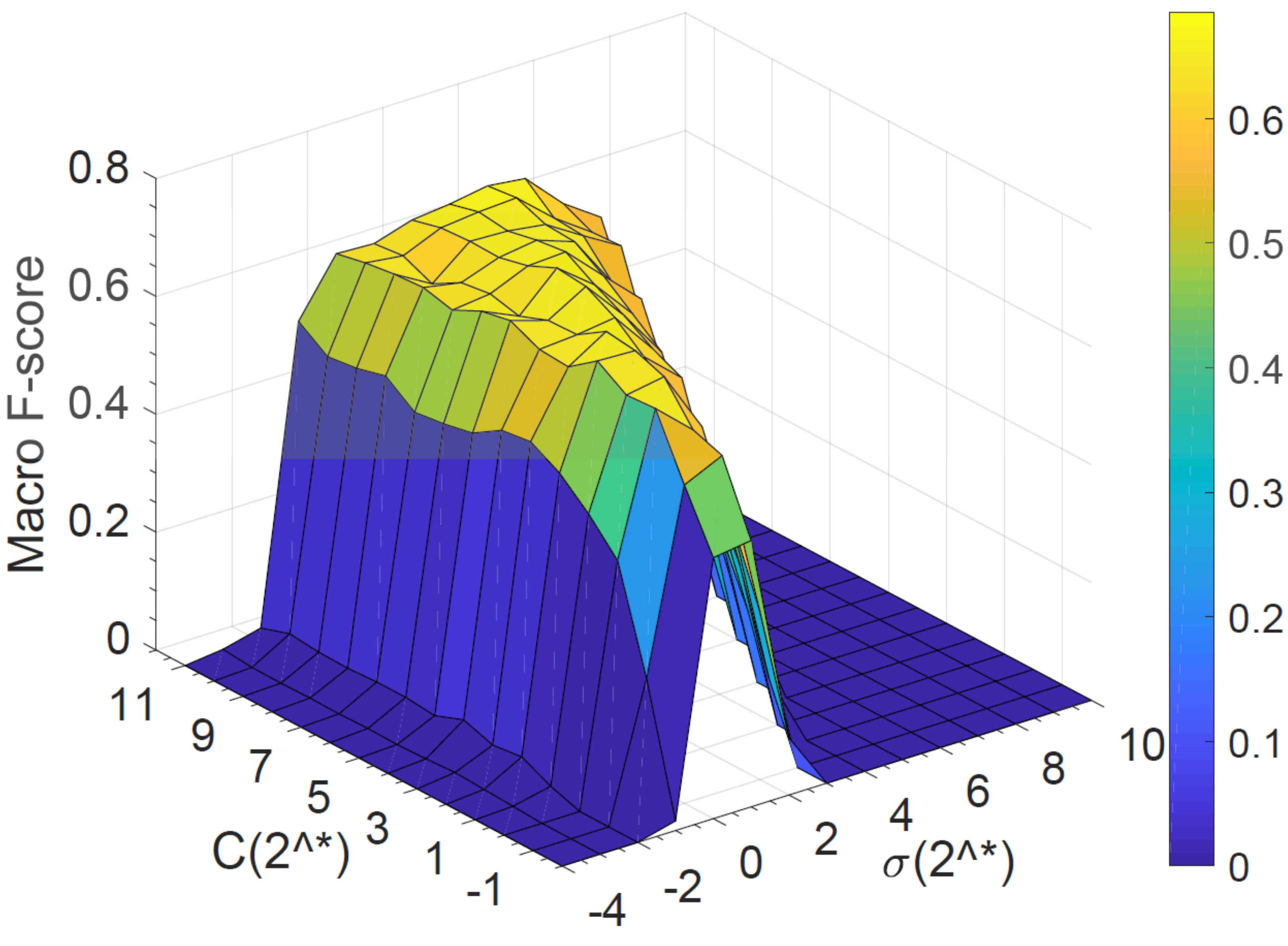}}\quad
\subfloat[${\rm MicF}$]{\includegraphics[width=2.5in]{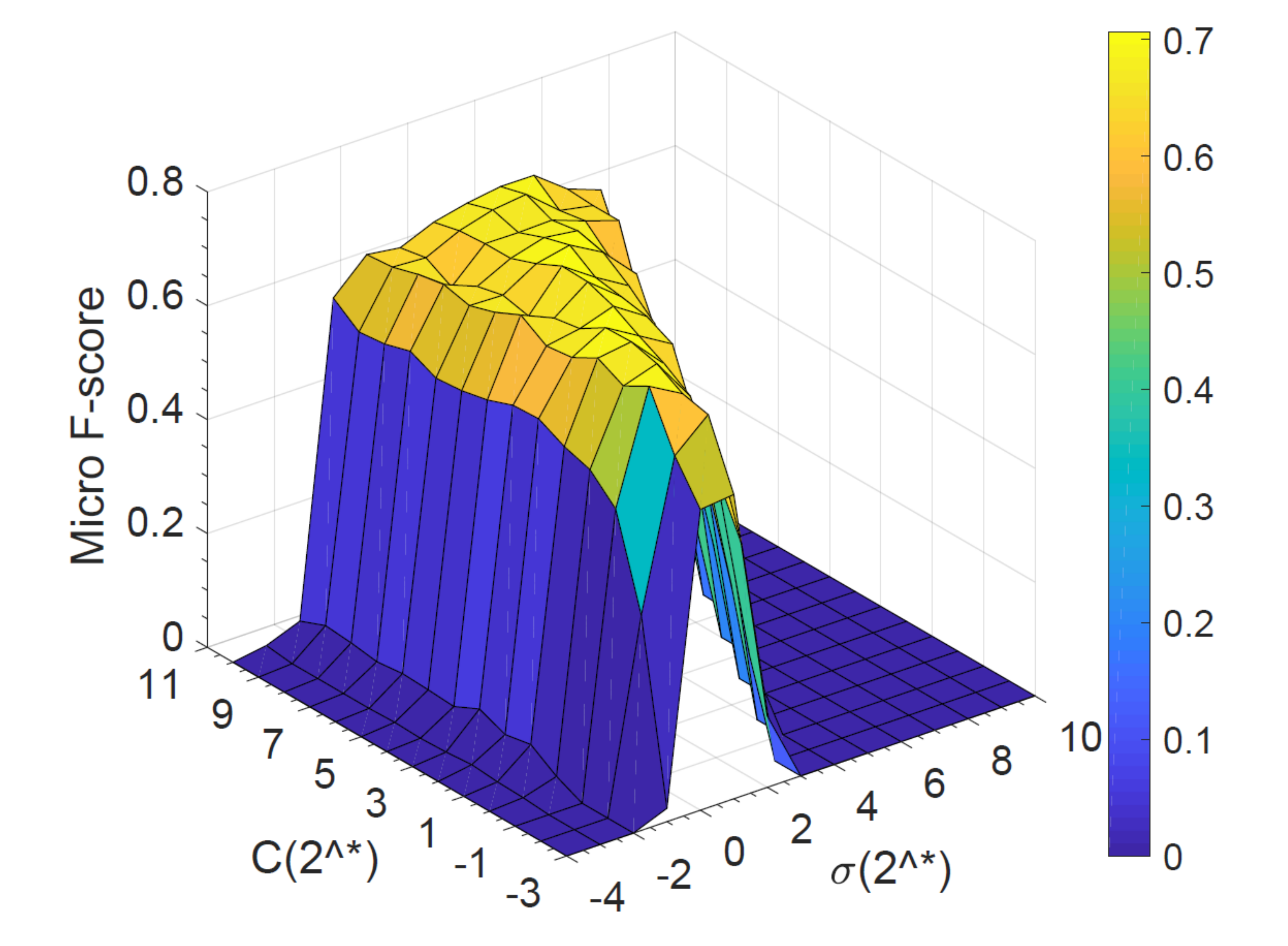}}
\caption{Sensitivity of SVM parameters regarding different metrics in {\sf BNCC} on {\it Emotions}.}
\label{fig.sensitivity}
\end{figure*}

The base classifiers in all the compared methods are binary, thus support vector machine (SVM)~\cite{vapnik2000nature,wang2013vector,wang2012inconsistency} is used consistently, since it has been verified to be ideal binary model in many applications. Furthermore, non-linear SVM is realized by applying RBF kernel $\mathcal{K}(\mathbf{x},\mathbf{x}_i)=\exp(-{\frac{||\mathbf{x}-\mathbf{x}_i||^2}{2\sigma^2}})$.

Since the performance of SVM is largely influenced by the model parameters, i.e., slack variable $C$ and kernel parameter $\sigma$, we first make an investigation on a small data set {\it Emotions}. We investigate the influences of $C$ and $\sigma$ on the performance of {\sf BNCC} by tuning $C$ from [$2^{-3},2^{-2},\dots,2^{9},2^{10}$] and $\sigma$ from [$2^{-3},2^{-2},\dots,2^{8},2^{9}$], where the result is demonstrated in Figure~\ref{fig.sensitivity}. It can be observed that the four metrics are very sensitive to $\sigma$ while are relatively stable regarding $C$, and this observation can also be found for the other methods. Usually, a larger value of $C$ can help improve the performance but will increase the execution time. Thus, in order to reduce validation time, we fix $C=100$ and tune parameter $\sigma$ for each data set. We conduct 10-fold cross-validation on the training set by tuning $\sigma=[2^{-3},2^{-2},\dots,2^{8},2^{9}]$, and the value that achieves the best validation performance is selected.

To avoid random effect and get comprehensive results, we perform 10-fold cross-validation with the selected parameter on the entire data set. This process is repeated for 10 times. Finally, the average result and standard deviation are reported. The experiments are conducted under MATLAB R2017b with the libsvm toolbox, which are executed on a computer with an Intel(R) Xeon(R) E5-2690 v3@2.60-GHz Core 2 Duo CPU, a 256-GB memory, and 64-bit Windows server system.

\subsection{Result Analysis}

\begin{table*}[!htbp]
\caption{Experimental Results of the Compared Methods on the Selected Data Sets Regarding Four Metrics}
\vspace{-0.5cm}
\begin{center}
\scalebox{0.45}{
\begin{tabular}{p{4.0cm}p{3.8cm}p{3.8cm}p{3.8cm}p{3.8cm}p{3.8cm}p{3.8cm}}
\hline
Data Sets &BR   &CLR  &CC   &GCC   &ECC   &BNCC \\
\hline\hline
\bf{HammingLoss}\\
Bibtex	
&0.0130 $\pm$ 0.0003&0.0139 $\pm$ 0.0002&0.0129 $\pm$ 0.0002&\bf{0.0119} $\pm$ \bf{0.0003}&0.0125 $\pm$ 0.0003&0.0129 $\pm$ 0.0001\\
CHD	
&0.2755 $\pm$ 0.0076&0.2799 $\pm$ 0.0075&0.2781 $\pm$ 0.0103&0.2853 $\pm$ 0.0279&0.2740 $\pm$ 0.0071&\bf{0.2712} $\pm$ \bf{0.0066}$\nearrow$\\
Education	
&0.0403 $\pm$ 0.0006&0.0422 $\pm$ 0.0008&0.0430 $\pm$ 0.0007&0.0422 $\pm$ 0.0025&\bf{0.0390} $\pm$ \bf{0.0005}&0.0430 $\pm$ 0.0012\\
Emotions
&0.1858 $\pm$ 0.0119&0.1925 $\pm$ 0.0067&0.1931 $\pm$ 0.0112&0.3779 $\pm$ 0.0446&\bf{0.1807} $\pm$ \bf{0.0085}&0.1904 $\pm$ 0.0153$\nearrow$\\
Enron	
&0.0498 $\pm$ 0.0009&0.0520 $\pm$ 0.0011&\bf{0.0496} $\pm$ \bf{0.0011}&0.0500 $\pm$ 0.0035&0.0509 $\pm$ 0.0012&0.0497 $\pm$ 0.0009\\
Flags
&0.2677 $\pm$ 0.0125&0.2636 $\pm$ 0.0172&0.2776 $\pm$ 0.0242&0.3567 $\pm$ 0.0347&0.2681 $\pm$ 0.0100&\bf{0.2621} $\pm$ \bf{0.0155}$\nearrow$\\
Genbase	
&0.0008 $\pm$ 0.0002&0.0310 $\pm$ 0.0031&0.0008 $\pm$ 0.0002&0.0012 $\pm$ 0.0009&\bf{0.0006} $\pm$ \bf{0.0002}&\bf{0.0006} $\pm$ \bf{0.0006}$\nearrow$\\
GramNegative
&0.0126 $\pm$ 0.0015&0.0251 $\pm$ 0.0050&\bf{0.0124} $\pm$ \bf{0.0020}&0.0149 $\pm$ 0.0039&0.0126 $\pm$ 0.0011&0.0126 $\pm$ 0.0010\\
GramPositive
&0.0289 $\pm$ 0.0058&0.0296 $\pm$ 0.0067&0.0318 $\pm$ 0.0072&0.0386 $\pm$ 0.0178&0.0317 $\pm$ 0.0043&\bf{0.0283} $\pm$ \bf{0.0053}$\nearrow$\\
HumanGoB3106
&0.0385 $\pm$ 0.0009&\bf{0.0375} $\pm$ \bf{0.0011}&0.0379 $\pm$ 0.0013&0.0441 $\pm$ 0.0032&0.0381 $\pm$ 0.0008&0.0390 $\pm$ 0.0012\\
HumanPseACC3106
&0.0659 $\pm$ 0.0012&0.0993 $\pm$ 0.0015&0.0674 $\pm$ 0.0021&0.0665 $\pm$ 0.0048&0.0683 $\pm$ 0.0018&\bf{0.0644} $\pm$ \bf{0.0020}$\nearrow$\\
Mediamill
&\bf{0.0299} $\pm$ \bf{0.0003}&0.0423 $\pm$ 0.0002&0.0304 $\pm$ 0.0006&0.0361 $\pm$ 0.0020&0.0300 $\pm$ 0.0005&0.0304 $\pm$ 0.0008\\
Medical	
&0.0095 $\pm$ 0.0007&0.0250 $\pm$ 0.0014&0.0093 $\pm$ 0.0009&0.0112 $\pm$ 0.0017&\bf{0.0091} $\pm$ \bf{0.0007}&0.0092 $\pm$ 0.0006$\nearrow$\\
PlanGoB978
&0.0391 $\pm$ 0.0020&0.0371 $\pm$ 0.0027&0.0392 $\pm$ 0.0025&\bf{0.0367} $\pm$ \bf{0.0054}&0.0372 $\pm$ 0.0022&0.0371 $\pm$ 0.0024$\nearrow$\\
Scene
&0.0699 $\pm$ 0.0027&0.0707 $\pm$ 0.0026&0.0730 $\pm$ 0.0039&0.1068 $\pm$ 0.0094&\bf{0.0693} $\pm$ \bf{0.0023}&0.0696 $\pm$ 0.0020$\nearrow$\\
Virus
&0.0433 $\pm$ 0.0072&0.0438 $\pm$ 0.0102&0.0392 $\pm$ 0.0075&0.0546 $\pm$ 0.0231&\bf{0.0383} $\pm$ \bf{0.0090}&0.0402 $\pm$ 0.0064\\
WaterQuality
&\bf{0.2851} $\pm$ \bf{0.0054}&0.3034 $\pm$ 0.0033&0.3461 $\pm$ 0.0094&0.3006 $\pm$ 0.0116&0.3047 $\pm$ 0.0057&0.3369 $\pm$ 0.0102$\nearrow$\\
Yeast	
&0.1964 $\pm$ 0.0033&0.1993 $\pm$ 0.0039&0.1949 $\pm$ 0.0028&0.2071 $\pm$ 0.0086&0.1934 $\pm$ 0.0030&\bf{0.1914} $\pm$ \bf{0.0040}$\nearrow$\\
\hline\hline
\bf{Fscore}\\
Bibtex
&0.4333 $\pm$ 0.0073&0.1971 $\pm$ 0.0081&0.4277 $\pm$ 0.0070&0.4227 $\pm$ 0.0124&\bf{0.4339} $\pm$ \bf{0.0121}&0.4306 $\pm$ 0.0091$\nearrow$\\
CHD
&0.6912 $\pm$ 0.0122&0.6764 $\pm$ 0.0066&0.6869 $\pm$ 0.0153&0.6457 $\pm$ 0.0425&0.6910 $\pm$ 0.0123&\bf{0.6945} $\pm$ \bf{0.0090}$\nearrow$\\
Education
&0.3873 $\pm$ 0.0077&0.1416 $\pm$ 0.0265&0.4050 $\pm$ 0.0086&0.3953 $\pm$ 0.0270&0.3972 $\pm$ 0.0082&\bf{0.4166} $\pm$ \bf{0.0124}$\nearrow$\\
Emotions
&0.6763 $\pm$ 0.0208&0.6437 $\pm$ 0.0189&0.6751 $\pm$ 0.0199&0.3699 $\pm$ 0.0586&0.6865 $\pm$ 0.0185&\bf{0.6928} $\pm$ \bf{0.0182}$\nearrow$\\
Enron
&0.5677 $\pm$ 0.0067&0.4936 $\pm$ 0.0239&0.5710 $\pm$ 0.0082&0.5266 $\pm$ 0.0397&0.5681 $\pm$ 0.0107&\bf{0.5733} $\pm$ \bf{0.0087}$\nearrow$\\
Flags
&0.7098 $\pm$ 0.0155&0.6985 $\pm$ 0.0205&0.6889 $\pm$ 0.0296&0.6160 $\pm$ 0.0566&0.6953 $\pm$ 0.0117&\bf{0.7104} $\pm$ \bf{0.0162}$\nearrow$\\
Genbase
&0.9930 $\pm$ 0.0029&0.3138 $\pm$ 0.0712&0.9930 $\pm$ 0.0030&0.9901 $\pm$ 0.0088&0.9948 $\pm$ 0.0027&\bf{0.9963} $\pm$ \bf{0.0038}$\nearrow$\\
GramNegative
&\bf{0.9630} $\pm$ \bf{0.0047}&0.8543 $\pm$ 0.0416&0.9611 $\pm$ 0.0071&0.9466 $\pm$ 0.0150&0.9566 $\pm$ 0.0044&0.9569 $\pm$ 0.0047\\
GramPositive
&0.9305 $\pm$ 0.0162&0.9284 $\pm$ 0.0142&0.9363 $\pm$ 0.0143&0.9263 $\pm$ 0.0345&0.9344 $\pm$ 0.0088&\bf{0.9432} $\pm$ \bf{0.0122}$\nearrow$\\
HumanGoB3106
&0.7932 $\pm$ 0.0049&0.7868 $\pm$ 0.0063&0.7982 $\pm$ 0.0093&0.7302 $\pm$ 0.0217&\bf{0.7969} $\pm$ \bf{0.0041}&0.7931 $\pm$ 0.0073\\
HumanPseACC3106
&0.6270 $\pm$ 0.0099&0.0050 $\pm$ 0.0010&0.6338 $\pm$ 0.0129&0.6110 $\pm$ 0.0332&0.6252 $\pm$ 0.0118&\bf{0.6384} $\pm$ \bf{0.0145}$\nearrow$\\
Mediamill
&\bf{0.5596} $\pm$ \bf{0.0022}&0.0178 $\pm$ 0.0041&0.5521 $\pm$ 0.0047&0.3635 $\pm$ 0.0811&0.5533 $\pm$ 0.0055&0.5542 $\pm$ 0.0121$\nearrow$\\
Medical
&0.8067 $\pm$ 0.0122&0.1180 $\pm$ 0.0660&0.8141 $\pm$ 0.0183&0.7771 $\pm$ 0.0347&0.8160 $\pm$ 0.0146&\bf{0.8238} $\pm$ \bf{0.0103}$\nearrow$\\
PlanGoB978
&0.7696 $\pm$ 0.0100&0.7685 $\pm$ 0.0160&0.7772 $\pm$ 0.0148&0.7746 $\pm$ 0.0340&\bf{0.7862} $\pm$ \bf{0.0129}&0.7858 $\pm$ 0.0123$\nearrow$\\
Scene
&0.7602 $\pm$ 0.0087&0.7467 $\pm$ 0.0092&0.7750 $\pm$ 0.0124&0.6939 $\pm$ 0.0275&0.7641 $\pm$ 0.0084&\bf{0.7844} $\pm$ \bf{0.0084}$\nearrow$\\
Virus
&0.9078 $\pm$ 0.0235&0.8950 $\pm$ 0.0261&0.9138 $\pm$ 0.0180&0.8707 $\pm$ 0.0554&0.9129 $\pm$ 0.0198&\bf{0.9143} $\pm$ \bf{0.0135}$\nearrow$\\
WaterQuality
&\bf{0.5517} $\pm$ \bf{0.0102}&0.5450 $\pm$ 0.0055&0.4297 $\pm$ 0.0275&0.4521 $\pm$ 0.0276&0.4348 $\pm$ 0.0119&0.4614 $\pm$ 0.0153$\nearrow$\\
Yeast
&0.6448 $\pm$ 0.0054&0.6147 $\pm$ 0.0102&0.6483 $\pm$ 0.0035&0.6284 $\pm$ 0.0195&0.6485 $\pm$ 0.0087&\bf{0.6612} $\pm$ \bf{0.0070}$\nearrow$\\
\hline\hline
\bf{MacF} \\
Bibtex
&0.2938 $\pm$ 0.0076&0.1343 $\pm$ 0.0086&0.2804 $\pm$ 0.0079&\bf{0.3236} $\pm$ \bf{0.0130}&0.2892 $\pm$ 0.0093&0.2863 $\pm$ 0.0060$\nearrow$\\
CHD
&0.5207 $\pm$ 0.0093&0.5039 $\pm$ 0.0084&0.5140 $\pm$ 0.0127&0.4789 $\pm$ 0.0609&0.5204 $\pm$ 0.0098&\bf{0.5264} $\pm$ \bf{0.0089}$\nearrow$\\
Education
&\bf{0.2056} $\pm$ \bf{0.0120}&0.1266 $\pm$ 0.0292&0.1978 $\pm$ 0.0134&0.1980 $\pm$ 0.0294&0.1980 $\pm$ 0.0147&0.1771 $\pm$ 0.0172\\
Emotions
&0.6638 $\pm$ 0.0188&0.6457 $\pm$ 0.0186&0.6486 $\pm$ 0.0223&0.3362 $\pm$ 0.0659&0.6633 $\pm$ 0.0241&\bf{0.6777} $\pm$ \bf{0.0297}$\nearrow$\\
Enron
&0.1784 $\pm$ 0.0092&0.1803 $\pm$ 0.0151&0.1751 $\pm$ 0.0062&0.1344 $\pm$ 0.0186&\bf{0.1985} $\pm$ \bf{0.0097}&0.1729 $\pm$ 0.0082\\
Flags
&0.6188 $\pm$ 0.0282&0.6193 $\pm$ 0.0327&0.6115 $\pm$ 0.0276&0.3713 $\pm$ 0.0655&0.6185 $\pm$ 0.0252&\bf{0.6358} $\pm$ \bf{0.0328}$\nearrow$\\
Genbase
&0.6615 $\pm$ 0.0393&0.2493 $\pm$ 0.0543&0.6631 $\pm$ 0.0203&0.6371 $\pm$ 0.0729&0.6572 $\pm$ 0.0230&\bf{0.6800} $\pm$ \bf{0.0518}$\nearrow$\\
GramNegative
&0.8046 $\pm$ 0.0296&0.7554 $\pm$ 0.0448&0.8062 $\pm$ 0.0352&0.8003 $\pm$ 0.0632&0.8472 $\pm$ 0.0345&\bf{0.8558} $\pm$ \bf{0.0308}$\nearrow$\\
GramPositive
&0.7521 $\pm$ 0.0406&0.7586 $\pm$ 0.0602&\bf{0.7905} $\pm$ \bf{0.0497}&0.7294 $\pm$ 0.0777&0.7729 $\pm$ 0.0373&0.7689 $\pm$ 0.0335\\
HumanGoB3106
&\bf{0.6464} $\pm$ \bf{0.0176}&0.6348 $\pm$ 0.0183&0.6417 $\pm$ 0.0217&0.6205 $\pm$ 0.0429&0.6373 $\pm$ 0.0222&0.6383 $\pm$ 0.0193\\
HumanPseACC3106
&0.0605 $\pm$ 0.0007&0.0014 $\pm$ 0.0006&\bf{0.0798} $\pm$ \bf{0.0034}&0.0748 $\pm$ 0.0121&0.0792 $\pm$ 0.0043&0.0611 $\pm$ 0.0008\\
Mediamill
&0.0289 $\pm$ 0.0001&0.0164 $\pm$ 0.0042&0.0296 $\pm$ 0.0008&0.0227 $\pm$ 0.0042&0.0284 $\pm$ 0.0002&\bf{0.0308} $\pm$ \bf{0.0020}$\nearrow$\\
Medical
&0.3406 $\pm$ 0.0090&0.0763 $\pm$ 0.0412&0.3157 $\pm$ 0.0084&0.3102 $\pm$ 0.0298&\bf{0.3470} $\pm$ \bf{0.0161}&0.3355 $\pm$ 0.0114$\nearrow$\\
PlanGoB978
&0.6860 $\pm$ 0.0266&0.6908 $\pm$ 0.0326&0.6684 $\pm$ 0.0246&0.6895 $\pm$ 0.0669&0.6730 $\pm$ 0.0251&\bf{0.7018} $\pm$ \bf{0.0281}$\nearrow$\\
Scene
&0.8001 $\pm$ 0.0073&0.7955 $\pm$ 0.0068&0.7938 $\pm$ 0.0111&0.7013 $\pm$ 0.0250&0.7991 $\pm$ 0.0068&\bf{0.8038} $\pm$ \bf{0.0077}$\nearrow$\\
Virus
&0.7801 $\pm$ 0.0601&0.7969 $\pm$ 0.0596&0.7954 $\pm$ 0.0382&0.7425 $\pm$ 0.1305&0.8001 $\pm$ 0.0779&\bf{0.8120} $\pm$ \bf{0.0269}$\nearrow$\\
WaterQuality
&0.4934 $\pm$ 0.0114&\bf{0.4966} $\pm$ \bf{0.0049}&0.3484 $\pm$ 0.0230&0.3695 $\pm$ 0.0273&0.2690 $\pm$ 0.0075&0.3803 $\pm$ 0.0188$\nearrow$\\
Yeast
&0.3287 $\pm$ 0.0042&0.3047 $\pm$ 0.0052&0.3452 $\pm$ 0.0058&\bf{0.3836} $\pm$ \bf{0.0198}&0.3379 $\pm$ 0.0055&0.3641 $\pm$ 0.0044$\nearrow$\\
\hline\hline
\bf{MicF} \\
Bibtex
&0.4393 $\pm$ 0.0074&0.2244 $\pm$ 0.0107&0.4343 $\pm$ 0.0040&\bf{0.4703} $\pm$ \bf{0.0133}&0.4464 $\pm$ 0.0120&0.4389 $\pm$ 0.0064$\nearrow$\\
CHD
&0.6894 $\pm$ 0.0102&0.6765 $\pm$ 0.0068&0.6844 $\pm$ 0.0136&0.6692 $\pm$ 0.0421&0.6914 $\pm$ 0.0081&\bf{0.6937} $\pm$ \bf{0.0074}$\nearrow$\\
Education
&0.4318 $\pm$ 0.0076&0.2069 $\pm$ 0.0365&0.4308 $\pm$ 0.0072&0.4218 $\pm$ 0.0222&\bf{0.4425} $\pm$ \bf{0.0082}&0.4327 $\pm$ 0.0143$\nearrow$\\
Emotions
&0.6829 $\pm$ 0.0188&0.6659 $\pm$ 0.0134&0.6746 $\pm$ 0.0181&0.3900 $\pm$ 0.0613&0.6932 $\pm$ 0.0195&\bf{0.6935} $\pm$ \bf{0.0218}$\nearrow$\\
Enron
&0.5529 $\pm$ 0.0058&0.4903 $\pm$ 0.0166&0.5537 $\pm$ 0.0074&0.5290 $\pm$ 0.0381&0.5529 $\pm$ 0.0089&\bf{0.5564} $\pm$ \bf{0.0068}$\nearrow$\\
Flags
&0.7276 $\pm$ 0.0151&0.7228 $\pm$ 0.0167&0.7142 $\pm$ 0.0229&0.6243 $\pm$ 0.0556&0.7176 $\pm$ 0.0131&\bf{0.7343} $\pm$ \bf{0.0165}$\nearrow$\\
Genbase
&0.9913 $\pm$ 0.0024&0.4637 $\pm$ 0.0886&0.9914 $\pm$ 0.0025&0.9868 $\pm$ 0.0094&0.9934 $\pm$ 0.0024&\bf{0.9941} $\pm$ \bf{0.0057}$\nearrow$\\
GramNegative
&0.9521 $\pm$ 0.0056&0.8957 $\pm$ 0.0255&\bf{0.9528} $\pm$ \bf{0.0076}&0.9424 $\pm$ 0.0151&0.9517 $\pm$ 0.0043&0.9519 $\pm$ 0.0038\\
GramPositive
&0.9423 $\pm$ 0.0117&0.9408 $\pm$ 0.0133&0.9367 $\pm$ 0.0142&0.9237 $\pm$ 0.0349&0.9367 $\pm$ 0.0084&\bf{0.9436} $\pm$ \bf{0.0107}$\nearrow$\\
HumanGoB3106
&0.7763 $\pm$ 0.0057&0.7788 $\pm$ 0.0062&0.7768 $\pm$ 0.0089&0.7243 $\pm$ 0.0204&\bf{0.7769} $\pm$ \bf{0.0049}&0.7720 $\pm$ 0.0062\\
HumanPseACC3106
&0.6144 $\pm$ 0.0083&0.0090 $\pm$ 0.0026&0.6195 $\pm$ 0.0117&0.6184 $\pm$ 0.0287&0.6125 $\pm$ 0.0103&\bf{0.6239} $\pm$ \bf{0.0116}$\nearrow$\\
Mediamill
&\bf{0.5297} $\pm$ \bf{0.0033}&0.0269 $\pm$ 0.0063&0.5213 $\pm$ 0.0057&0.3834 $\pm$ 0.0752&0.5238 $\pm$ 0.0050&0.5246 $\pm$ 0.0122$\nearrow$\\
Medical
&0.8223 $\pm$ 0.0125&0.1728 $\pm$ 0.0866&0.8257 $\pm$ 0.0169&0.7904 $\pm$ 0.0314&0.8275 $\pm$ 0.0126&\bf{0.8293} $\pm$ \bf{0.0106}$\nearrow$\\
PlanGoB978
&0.7807 $\pm$ 0.0105&0.7890 $\pm$ 0.0157&0.7787 $\pm$ 0.0132&0.7874 $\pm$ 0.0300&0.7893 $\pm$ 0.0128&\bf{0.7905} $\pm$ \bf{0.0127}$\nearrow$\\
Scene
&0.7927 $\pm$ 0.0080&0.7876 $\pm$ 0.0077&0.7860 $\pm$ 0.0114&0.6928 $\pm$ 0.0265&0.7924 $\pm$ 0.0073&\bf{0.7960} $\pm$ \bf{0.0061}$\nearrow$\\
Virus
&0.8966 $\pm$ 0.0165&0.8923 $\pm$ 0.0244&\bf{0.9047} $\pm$ \bf{0.0185}&0.8628 $\pm$ 0.0557&0.9034 $\pm$ 0.0223&0.9039 $\pm$ 0.0135\\
WaterQuality
&\bf{0.5584} $\pm$ \bf{0.0095}&0.5476 $\pm$ 0.0047&0.4326 $\pm$ 0.0260&0.4781 $\pm$ 0.0236&0.4436 $\pm$ 0.0084&0.4696 $\pm$ 0.0159$\nearrow$\\
Yeast
&0.6391 $\pm$ 0.0062&0.6131 $\pm$ 0.0094&0.6446 $\pm$ 0.0044&0.6451 $\pm$ 0.0171&0.6435 $\pm$ 0.0064&\bf{0.6570} $\pm$ \bf{0.0066}$\nearrow$\\
\hline\noalign{\smallskip}
\end{tabular}}
\begin{minipage}{14cm}
\scriptsize{\textbf{Note: } {\it For each data set, the best performance is in bold face, and $\nearrow$ represents that the performance of BNCC is better than the original CC method.}}
\end{minipage}
\label{tab.results}
\end{center}
\end{table*}

Table~\ref{tab.results} reports the experimental results of the compared methods on the selected data sets in terms of the four evaluation metrics, where the best result for each data set is marked in bold. Basically, we have the following observations.

\begin{itemize}
\item All the methods have achieved low ${\rm HammingLoss}$ on most selected data sets, where {\sf BNCC} and {\sf ECC} are the best performing ones. However, as defined in Eq.~(\ref{eq.hammingLoss}), ${\rm HammingLoss}$ computes the ratio of the correctly identified instance-label pairs without distinguishing whether they are positive or negative, which means that the low ${\rm HammingLoss}$ might be induced by high recognition rate on negative labels regarding each instance, which are not important in MLL. In this case, ${\rm HammingLoss}$ is less valuable compared with the other three metrics. Thus, we put our focus on ${\rm Fscore}$, ${\rm MacF}$, and ${\rm MicF}$.

\item The proposed {\sf BNCC} has obtained satisfactory performance in most cases. It has achieved the best performance on most of the selected data sets regarding ${\rm Fscore}$ and ${\rm MicF}$, and half of the selected data sets regarding ${\rm MacF}$. The success of {\sf BNCC} is due to the employment of BN, which  makes fully use of the information in training data by Bayesian statistics and provides valuable information on how labels are correlated with each other in the whole system. Furthermore, we have taken into account both positive relationships and negative relationships in correlation analysis. If a label is positively or negatively correlated to another label, we can encourage its prediction result to be more similar or dissimilar to that of the other label.

\item Among the six compared methods, {\sf BR} is a first-order strategy that learns each label independently; {\sf CLR} is a second-order strategy that considers the pairwise correlations between two labels; {\sf CC}, {\sf GCC}, {\sf ECC}, and {\sf BNCC} are high-order strategies that explore the complex correlations among multiple labels. It can be observed that the proposed {\sf BNCC} can overall outperform the basic {\sf BR} and {\sf CLR}. However, there are exceptions like {\it Mediamill}, i.e., {\sf BR} presents very competitive performance regarding ${\rm Fscore}$ and ${\rm MicF}$. We speculate that this is because the labels in such data sets possess very weak correlation.

\item It is investigated that {\sf GCC} has obtained better results on few data sets than {\sf BNCC}. This may due to the fact that such data sets have strong local correlations, and capturing correlations locally is more effective than exploring them in a global system. Moreover, {\sf ECC} has also obtained better results on few data sets than {\sf BNCC}, this is because {\sf ECC} applies the ensemble mechanism, which overcomes the default of the random mechanism and largely improves the robustness and stability of traditional CC.
\end{itemize}
In summary, the proposed {\sf BNCC} is competitive compared with the other MLL methods regarding ${\rm Fscore}$, ${\rm MacF}$, and ${\rm MicF}$.

\begin{table}[!htbp]
\caption{Execution Time of the Compared Methods (Seconds)}
\begin{center}
\scalebox{0.65}{
\begin{tabular}{p{4.0cm}p{2.5cm}p{2.5cm}p{2.5cm}p{2.5cm}p{2.5cm}p{2.0cm}}
\hline
Data Sets &BR   &CLR  &CC   &GCC   &ECC   &BNCC \\
\hline\hline
\bf{Training Time}\\
Bibtex  	    &622.3 	   &3105.5 	    &5469.4   &37520.5 	  &31257.3 	    &16429.7\\
CHD	            &0.3 	   &0.4 	    &0.7 	  &3.1 	      &3.2 	        &1.1\\
Education	    &60.1 	   &88.0 	    &379.8 	  &2933.3 	  &2832.7 	    &354.4\\
Emotions	    &0.3 	   &0.5 	    &0.5 	  &3.2 	      &5.6 	        &0.9\\
Enron	        &50.4 	   &58.7      	&250.7 	  &681.1 	  &731.4      	&350.7\\
Flags	        &0.1 	   &0.1 	    &0.0 	  &0.8 	      &0.4       	&0.2\\
Genbase       	&0.5 	   &1.2 	    &2.6 	  &42.3       &22.7      	&8.5\\
GramNegative	&0.6 	   &0.9 	    &5.2 	  &245.4 	  &19.2      	&4.3\\
GramPositive	&0.1 	   &0.1 	    &0.4 	  &7.9 	      &3.0       	&1.0\\
HumanGoB3106	&14.2 	   &29.4 	    &296.9 	  &4987.5 	  &1471.8    	&323.6\\
HumanPseACC3106	&18.8 	   &16.6 	    &10.6 	  &55.0 	  &46.7 	    &10.0\\
Mediamill	    &1145.1    &2566.9      &3309.0   &12199.4 	  &26366.9    	&7216.4\\
Medical	        &2.0 	   &4.5 	    &12.2 	  &104.7 	  &139.1     	&60.3\\
PlanGoB978	    &1.1 	   &2.0         &10.9 	  &144.3 	  &54.7      	&11.8\\
Scene	        &13.5 	   &15.9 	    &36.3 	  &90.6 	  &457.4     	&25.4\\
Virus	        &0.1 	   &0.0      	&0.1 	  &1.6 	      &0.6 	        &0.3\\
WaterQuality	&1.6 	   &4.4       	&2.1 	  &19.5 	  &9.0 	        &4.0\\
Yeast	        &10.2 	   &19.7      	&24.1 	  &170.8 	  &403.7 	    &30.6\\
\hline\hline
\bf{Testing Time}\\
Bibtex	         &668.0    &19254.0 	&573.3 	 &535.9 	 &3125.8 	   &450.5\\
CHD	             &0.1 	   &0.1 	    &0.0 	 &0.0 	     &0.2 	       &0.1\\
Education	     &30.3 	   &117.7 	    &34.4 	 &30.5       &258.0 	   &21.6\\
Emotions	     &0.1 	   &0.1      	&0.0 	 &0.1 	     &0.4 	       &0.0\\
Enron	         &30.3 	   &92.0 	    &22.6 	 &7.3 	     &68.1 	       &22.3\\
Flags	         &0.1 	   &0.0      	&0.0 	 &0.0 	     &0.0 	       &0.0\\
Genbase	         &0.3 	   &3.1      	&0.2 	 &0.8 	     &1.9 	       &0.2\\
GramNegative  	 &0.6 	   &0.9      	&0.5 	 &3.3 	     &1.8 	       &0.4\\
GramPositive	 &0.1 	   &0.1 	    &0.0 	 &0.3 	     &0.3 	       &0.1\\
HumanGoB3106	 &31.7 	   &88.9 	    &31.2 	 &63.6	     &155.6 	   &33.2\\
HumanPseACC3106	 &3.7 	   &3.2 	    &0.6 	 &0.6 	     &3.1 	       &0.5\\
Mediamill	     &92.3     &431.6 	    &175.8 	 &47.2 	     &1370.0 	   &161.2\\
Medical	         &1.4 	   &8.9 	    &1.1 	 &1.9 	     &12.8 	       &1.1\\
PlanGoB978	     &1.1 	   &3.3 	    &1.1 	 &2.5 	     &5.4 	       &1.1\\
Scene	         &1.4 	   &2.7      	&2.1 	 &1.0 	     &32.1 	       &1.3\\
Virus	         &0.1 	   &0.0       	&0.0 	 &0.1 	     &0.1 	       &0.0\\
WaterQuality	 &0.1 	   &0.3         &0.1 	 &0.2 	     &0.6 	       &0.1\\
Yeast	         &2.0 	   &6.3 	    &1.3 	 &1.3 	     &25.1 	       &1.5\\
\hline\noalign{\smallskip}
\end{tabular}}
\label{tab.time}
\end{center}
\end{table}

Table~\ref{tab.time} reports the average training time and testing time of the six methods. It is investigated that regarding the training complexity, {\sf BR} is the fastest method, {\sf GCC} and {\sf ECC} are very time-consuming, while {\sf CLR}, {\sf CC}, and the proposed {\sf BNCC} are at an intermediate level. The high complexity of {\sf GCC} probably arises from the additional clustering process, and the high complexity of {\sf ECC} comes from the training process of the multiple CCs. Regarding the testing complexity, the most time-consuming methods are {\sf CLR} and {\sf ECC}, while the other four methods are much more efficient. In summary, we can get the conclusion that the proposed {\sf BNCC} can achieve satisfactory performance in a relatively low time complexity in both training and testing.

\begin{table}[!htbp]
\caption{Paired Wilcoxon's Signed-rank Tests}
\begin{center}
\scalebox{0.75}{
\begin{tabular}{p{4.0cm}p{2.5cm}p{2.5cm}p{2.5cm}p{2.5cm}p{2.0cm}}
\hline
Method &CLR  &CC   &GCC   &ECC   &BNCC  \\
\hline\hline
\bf{HammingLoss}\\
BR	        &0.0057$^\surd$   &0.1125           &0.0014$^\surd$   &0.3085 	        &0.3086\\
CLR        	&- 	              &0.2485          	&0.5861 	      &0.0065$^\surd$  	&0.0106$^\surd$\\
CC	        &- 	              &-	            &0.0294$^\surd$   &0.0089$^\surd$   &0.0094$^\surd$\\
GCC	        &- 	              &-                &-	              &0.0053$^\surd$   &0.0096$^\surd$\\
ECC	        &- 	              &- 	            &-	              &- 	            &0.8563\\

\hline\hline
\bf{Fscore}\\
BR	        &0.0002$^\surd$  &0.4348 	        &0.0006$^\surd$    &0.3163 	        &0.0311$^\surd$\\
CLR	        &-        	     &0.0021$^\surd$ 	&0.2485 	       &0.0016$^\surd$ 	&0.0012$^\surd$\\
CC	        &- 	             &-	                &0.0007$^\surd$    &0.6951 	        &0.0018$^\surd$\\
GCC	        &-	             &- 	            &- 	               &0.0007$^\surd$ 	&0.0002$^\surd$\\
ECC	        &-             	 &	                &-	               &- 	            &0.0025$^\surd$\\

\hline\hline
\bf{MacF}\\
BR	        &0.0108$^\surd$  &0.3271 	        &0.0139$^\surd$ 	&0.7112 	      &0.2311\\
CLR	        &-	             &0.0386$^\surd$ 	&0.9826 	        &0.0096$^\surd$   &0.0029$^\surd$\\
CC	        &- 	             &-	                &0.0346$^\surd$ 	&0.3061 	      &0.0429$^\surd$\\
GCC      	&-	             &-	                &-	                &0.0442$^\surd$   &0.0096$^\surd$\\
ECC         &-               &- 	            &-	                &-	              &0.1989\\

\hline\hline
\bf{MicF}\\
BR	        &0.0007$^\surd$  &0.1445 	        &0.0033$^\surd$ 	&0.3560 	        &0.0451$^\surd$\\
CLR	        &-	             &0.0156$^\surd$ 	&0.3061 	        &0.0057$^\surd$ 	&0.0021$^\surd$\\
CC	        &-	             &- 	            &0.0139$^\surd$ 	&0.0157$^\surd$ 	&0.0016$^\surd$\\
GCC         &- 	             &- 	            &- 	                &0.0079$^\surd$ 	&0.0021$^\surd$\\
ECC	        &- 	             &-	                &-                  &- 	                &0.0386$^\surd$\\

\hline\noalign{\smallskip}
\end{tabular}}
\begin{minipage}{15cm}
\scriptsize{\textbf{Note:} For each test, $\surd$ represents that the two compared methods are significantly different with $\alpha=0.05$.}
\end{minipage}
\label{tab.tests}
\end{center}
\end{table}

Finally, we make some statistical tests on the results listed in Table~\ref{tab.results}. Paired Wilcoxon's signed rank tests are performed, where the $p$-values are reported in Table~\ref{tab.tests}. It is a famous nonparametric statistical hypothesis test for assessing that whether there exists statistical difference between the results of two methods. If the $p$-value between two methods is smaller than the significance level $\alpha$, the two compared methods are regarded as statistically different, otherwise not. It can be seen that regarding ${\rm Fscore}$ and ${\rm MicF}$, {\sf BNCC} is statistically different from all of the other methods. However, regarding ${\rm MacF}$, {\sf BNCC} is not significantly better than {\sf ECC}. According to Table~\ref{tab.time}, we know that either the training and testing of {\sf ECC} is much more expensive than {\sf BNCC}, thus we can get the conclusion that {\sf BNCC} is a competitive method compared with {\sf ECC} with a much higher efficiency.

\section{Conclusions}\label{sec.conlusions}

This paper proposes a BNCC method by discovering and incorporating the correlations among labels. It employs conditional entropy to evaluate the uncertainty of deciding a label under the condition of other labels, such that the directed dependencies among labels can be modelled. By defining a new scoring function, a heuristic algorithm is proposed to learn the BN structure, and topological sorting is utilized to determine the final label order. Empirical studies show that the proposed BNCC method has achieved competitive performance compared with several state-of-the-art MLL approaches. Especially, it has improved the performance of traditional CC approach without increasing the time complexity in both training and testing. Future research regarding this topic may include the following directions: 1) in this paper, we only realize the BNCC method on SVM classifiers, it will be interesting to discuss the effectiveness of the method with regard to different base models; 2) other scoring functions for BN can be designed, which will lead to new evaluation measures and heuristic algorithms; 3) the label correlation analysis in this paper can be further applied to other MLL methods rather than CC approach; 4) given the multi-objective characteristics of the machine learning problem itself, it should be interesting to explore a multi-objective formulation of BNCC method. Current advancements in the evolutionary multi-objective optimization are worthwhile to be considered~\cite{LiFK11,LiKM11,CaoKWL12,LiKCLZS12,LiKWCR12,LiWKC13,LiKWTM13,LiK14,CaoKWL14,LiFKZ14,LiZKLW14,CaoKWLLK15,LiDZ15,LiKD15,LiDZK15,WuKZLWL15,LiKZD15,LiODY16,WuLKZZ17,WuKJLZ17,LiDZZ17,LiDAY17,LiDY18,LiCMY18,WuLKZ18,ChenLY18,ChenLBY18,LiCFY19,LiCSY19,WuLKZZ19}

\section*{Acknowledgement}

This work was supported in part by the National Natural Science Foundation of China (Grant 61772344, Grant 61732011, and Grant 61811530324), in part by the HD Video R\&D Platform for Intelligent Analysis and Processing in Guangdong Engineering Technology Research Centre of Colleges and Universities (Grant GCZX-A1409), in part by the Natural Science Foundation of Shenzhen (Grant JCYJ20170818091621856), in part by the Natural Science Foundation of SZU (Grant 827-000140, Grant 827-000230, and Grant 2017060), and in part by the Interdisciplinary Innovation Team of Shenzhen University. K. Li was supported by UKRI Future Leaders Fellowship (Grant No. MR/S017062/1) and Royal Society (Grant No. IEC/NSFC/170243)

\bibliographystyle{plain}
\bibliography{reference}

\end{document}